\documentclass[twocolumn,10pt,final]{asme2ej}

\usepackage{graphicx} 
\usepackage{hyperref}   
\usepackage{amsmath}
\hypersetup{
	colorlinks=true,
	linkcolor=blue,
	citecolor=blue,
	urlcolor=blue,
}
\usepackage[square,numbers]{natbib}
\usepackage{booktabs}
\usepackage{subcaption}
\usepackage{caption}
\usepackage{float}
\usepackage{array}

\usepackage[compact]{titlesec}   
\usepackage{enumitem}
\setlist{noitemsep, topsep=0pt, leftmargin=*}

\title{Continual Learning Strategies for 3D Engineering Regression Problems: A Benchmarking Study}

\author{Kaira M. Samuel
    \affiliation{
	Computational Science and Engineering\\
	Massachusetts Institute of Technology\\
	Cambridge, Massachusetts 02139\\
    Email: kmsamuel@mit.edu
    }	
}

\author{Faez Ahmed 
    \affiliation{ 
	Department of Mechanical Engineering\\
	Massachusetts Institute of Technology\\
	Cambridge, Massachusetts 02139\\
    Email: faez@mit.edu
    }	
}
\begin{document}

\maketitle    

\begin{abstract}

{\it Engineering problems that apply machine learning often involve computationally intensive methods but rely on limited datasets. As engineering data evolves with new designs and constraints, models must incorporate new knowledge over time. However, high computational costs make retraining models from scratch infeasible. Continual learning (CL) offers a promising solution by enabling models to learn from sequential data while mitigating catastrophic forgetting, where a model forgets previously learned mappings. This work introduces CL to engineering design by benchmarking several CL methods on representative regression tasks. We apply these strategies to five engineering datasets and construct nine new engineering CL benchmarks to evaluate their ability to address forgetting and improve generalization. Preliminary results show that applying existing CL methods to these tasks improves performance over naive baselines. In particular, the Replay strategy achieved performance comparable to retraining in several benchmarks while reducing training time by nearly half, demonstrating its potential for real-world engineering workflows. The code and datasets used in this work will be available at: \href{https://github.com/kmsamuel/cl-for-engineering-release}{https://github.com/kmsamuel/cl-for-engineering-release}.
\\}

\end{abstract}

\section{Introduction}


Machine learning (ML) can be leveraged in engineering applications to advance research in areas such as aerodynamic prediction, rapid prototyping, and design optimization. Traditionally, these tasks have relied on computationally expensive and time-intensive simulations, including computational fluid dynamics (CFD) simulations and finite element analysis (FEA). To address these limitations, ML methods have been developed to obtain field variable predictions more quickly by circumventing these time-consuming processes \cite{sun2019review}.

In particular, trained ML models can replace costly high-fidelity engineering simulations in a process called \textit{surrogate modeling}. In engineering applications, obtaining high-fidelity predictions of continuous values (i.e., drag coefficients, stresses, thermal loads) is essential, making regression a key focus. Such engineering regression tasks can span from predicting aerodynamic quantities, such as drag coefficients from car \cite{elrefaie2025drivaernet}\cite{song2023surrogate} or airplane geometries \cite{sabater2022fast}, to computing material properties of meta-material designs \cite{lee2024data}. In all of these applications, successful surrogate models define precise mappings between input features and continuous target values. However, this success is highly dependent on access to large, high-quality training datasets that cover the full design space, which are often expensive to obtain in engineering contexts.

Furthermore, data availability is rarely static in real-world design pipelines. Engineering design processes are inherently dynamic: as new data, constraints, or performance targets emerge, shifting data distributions, models must adapt accordingly. Numerous examples illustrate the importance of continual updates. A digital twin, for example, aims to mirror a physical system's state in real time and must ingest streaming data as the system changes \cite{wagg2020digital, jiang2021industrial}. Likewise, a surrogate model for rapid aerodynamics or structural prediction may need to expand its valid range whenever new design candidates or simulation regimes are introduced. This poses a challenge for traditional surrogate modeling approaches, which typically assume a fixed training set. Standard practice would require merging old and new data, and then retraining from scratch, which can be computationally expensive and prone to data management challenges \cite{verwimp2023continual}.

Here, \textit{ continual learning} offers a promising alternative by allowing models to update incrementally and learn from new data more efficiently than through retraining, mitigating \textit{catastrophic forgetting} (where the model's performance on old data degrades as it encounters new data). This stands in stark contrast to the conventional train-once-and-deploy paradigm, which is increasingly mismatched to the evolving nature of modern engineering workflows.

However, most continual learning research focuses on classification in domains like computer vision or NLP, leaving regression tasks and engineering problems under-explored. This paper explores how classical CL strategies--- such as rehearsal-based methods (Experience Replay, Gradient Episodic Memory) and parameter regularization (Elastic Weight Consolidation)--- perform in these challenging engineering regression scenarios. We motivate three types of task sequences that commonly arise and assemble a suite of engineering datasets to benchmark two of them. By systematically comparing CL approaches, we reveal best practices, limitations, and prospects for seamlessly integrating CL into real engineering workflows.

The main contributions of this paper are summarized as follows:
\begin{enumerate}
    \item Introduce continual learning to engineering regression tasks, specifically for surrogate modeling.
    \item Propose continual learning scenarios tailored explicitly for regression tasks: \textit{bin incremental}, \textit{input incremental}, and \textit{multi-target incremental}, aligning with realistic engineering data generation workflows.
    \item Benchmark several established continual learning strategies, including Experience Replay (ER), Elastic Weight Consolidation (EWC), and Gradient Episodic Memory (GEM), across five representative 3D engineering datasets and two regression continual learning scenarios, creating nine distinct benchmarks to facilitate systematic evaluation.
    \item Demonstrate empirically that adapting existing continual learning methods can significantly mitigate catastrophic forgetting and improve generalizability in engineering surrogate modeling tasks compared to naive incremental training.
    \item Analyze the effectiveness of each strategy, highlighting the impact of data representation, dataset size, and dimensionality on performance, thereby providing practical guidelines for the integration of continual learning in engineering design processes.
\end{enumerate}

\section{Preliminaries}
In this section, we discuss the empirical motivations and background knowledge relevant to the continual learning setting. We first provide an example of the problem on one of our engineering benchmarks and subsequently define important terminology in continual learning.

\begin{figure*}[!htb]
    \centering
    \begin{subfigure}{0.5\textwidth}
        \centering
        \includegraphics[width=\linewidth]{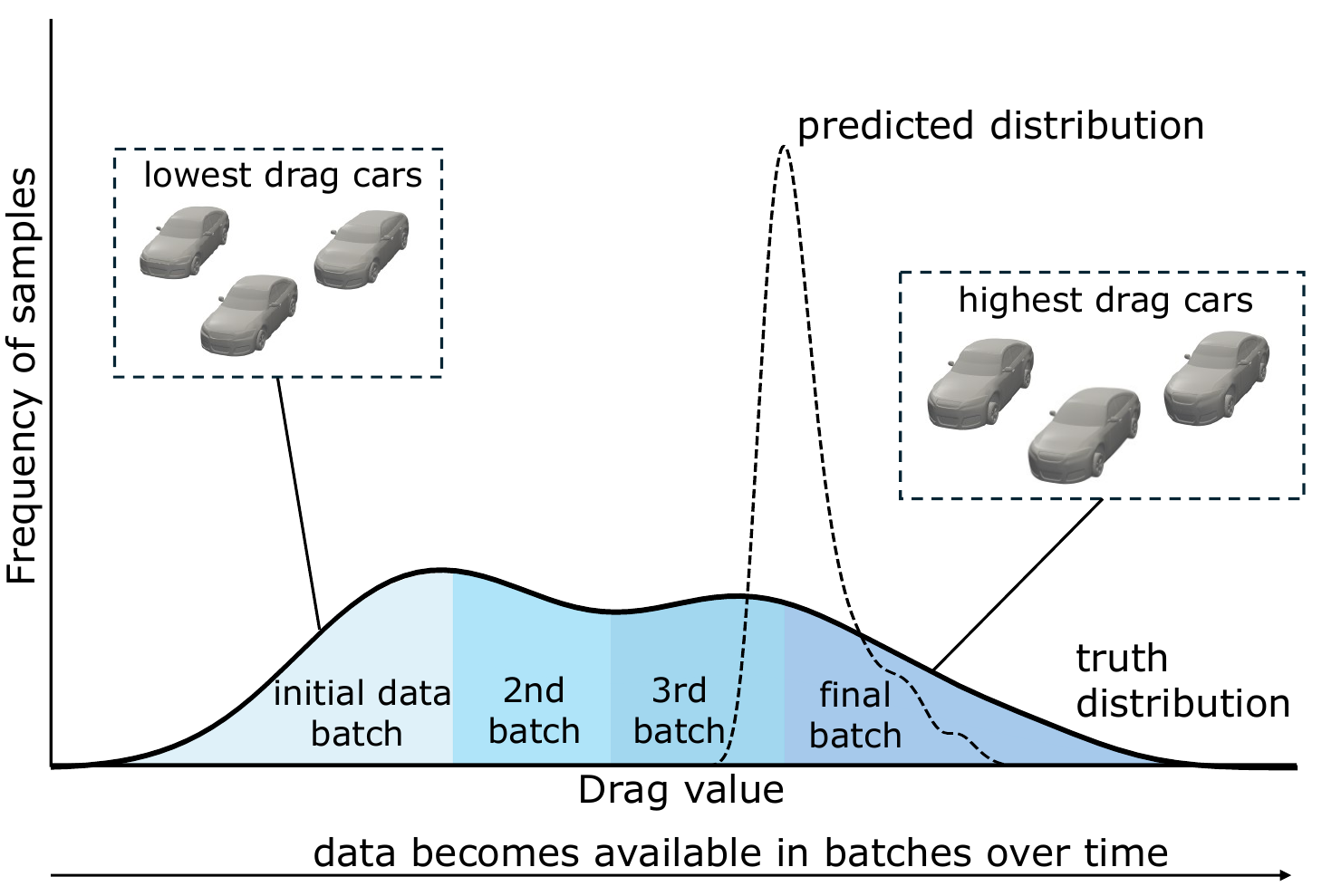}
        \caption{}
        \label{fig:forgetting_kdes}
    \end{subfigure}
    \hfill
    \begin{subfigure}{0.4\textwidth}
        \centering
        \includegraphics[width=\linewidth]{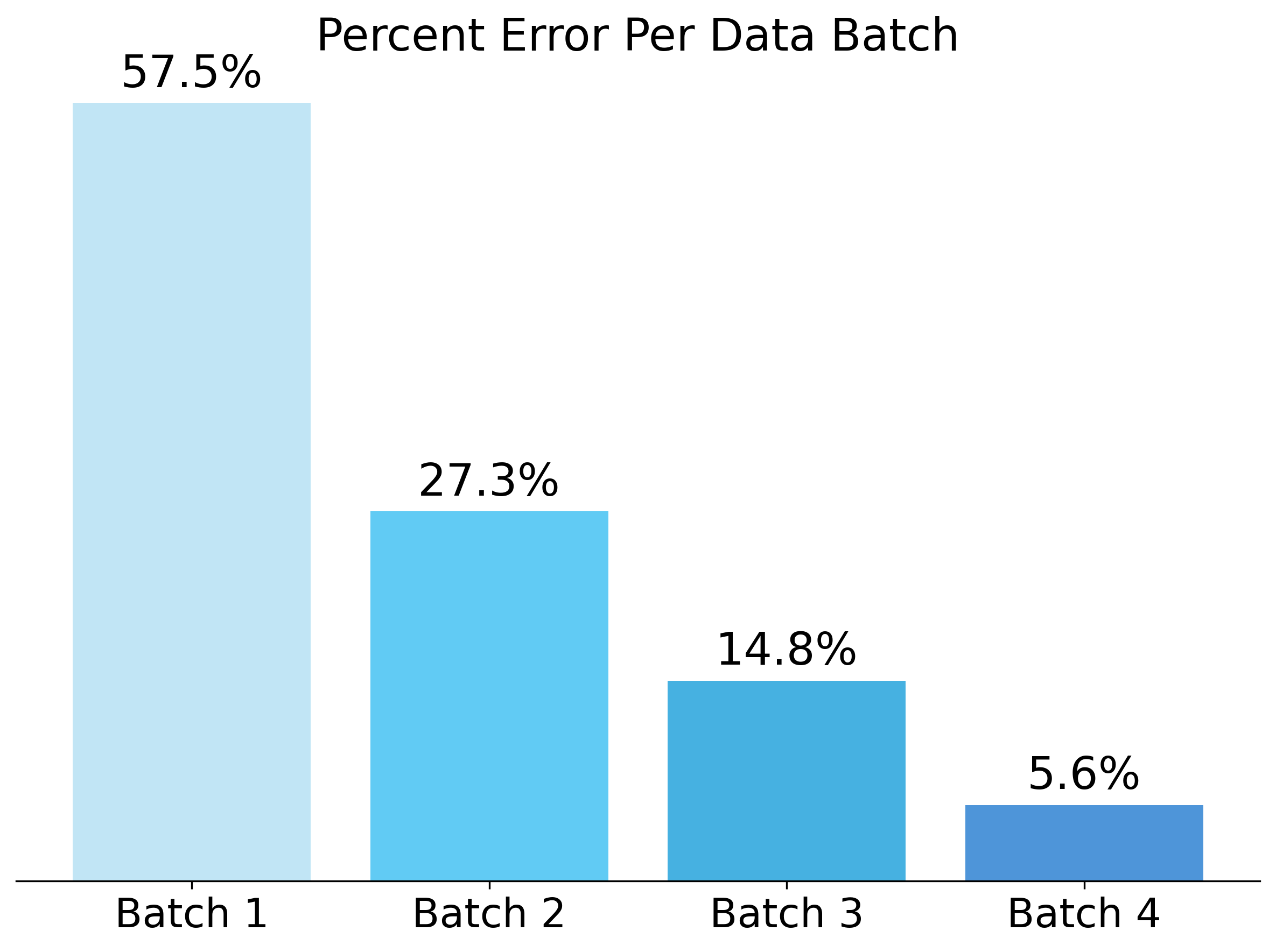}
        \caption{}
        \label{fig:forgetting_errors}
    \end{subfigure}
    \caption{Catastrophic forgetting is demonstrated using the car drag prediction problem from the DrivAerNet dataset. A surrogate model is trained on each data batch as the batches incrementally become available. Fig. \ref{fig:forgetting_kdes} shows a KDE plot of the final model's predictions on the entire dataset, which consists of all batches. As seen, the model exclusively predicts values within the third and fourth batches, demonstrating a lack of retention of earlier batch samples. Fig \ref{fig:forgetting_errors} emphasizes this discrepancy by reporting the percent error in predictions for each batch.}
    \label{fig:forgetting_demo}
\end{figure*}

\subsection*{Problem Demonstration}
The DrivAerNet dataset is used to demonstrate the problems associated with naively training a model on dynamic data (see Figure \ref{fig:forgetting_demo}). The dataset consists of geometric car geometries as inputs and their CFD-obtained coefficient of drag values as target values. In this scenario, the data becomes available in batches, which is an expected aspect of data generation cycles. In the extreme case, we can assume that the new batches of data do not overlap in their drag value distributions; thus, the new batches of data contain increasing drag values within their distribution range, depicted in Figure \ref{fig:forgetting_kdes}. A model is trained to predict the drag value from the input car geometry. It is trained on each batch incrementally to simulate new data becoming available at different times, and the final model is evaluated, with the kernel density estimate (KDE) distributions for the truth values (solid curve) and predicted values (perforated curve) plotted in Figure \ref{fig:forgetting_kdes}.

We observe that the earlier batches of data have a much higher percent error than the final batch (batch 4), as seen in Figure \ref{fig:forgetting_errors}. These results demonstrate the phenomenon of catastrophic forgetting, the key problem targeted by the domain of continual learning. Thus, we demonstrate a need for better suited strategies of retaining old distributions in the engineering context without requiring expensive retraining on all previous data.

\subsection*{What is Continual Learning?}
Continual learning---the ability to learn from a stream of tasks or data without forgetting past knowledge---has been extensively studied in the context of classification problems. In fact, numerous CL algorithms exist for image or text classification benchmarks  \cite{belouadah2021comprehensive, masana2022class, 726791, pmlr-v78-lomonaco17a}. However, applying CL to regression tasks (predicting continuous outputs) has been far less explored. As of 2021, He and Sick noted ``there is currently no research that addresses the catastrophic forgetting problem in regression tasks''~\cite{he2021clear}.

\begin{figure*}[h]
  \centering
  \includegraphics[width=\textwidth]{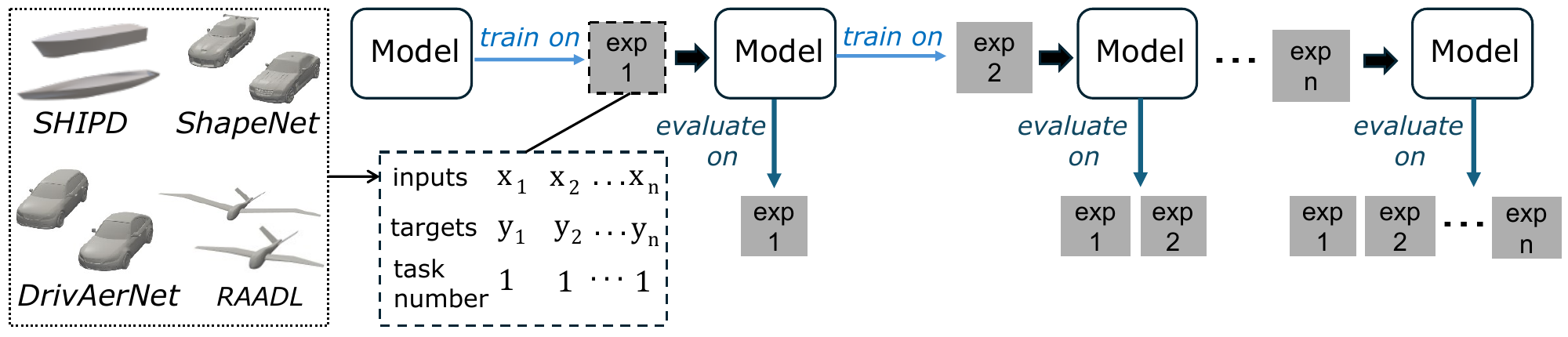}
  \caption{Continual learning model training scheme, in which a model is incrementally trained on new experiences and evaluated on all experiences seen up until that point.}
  \label{fig:CL_training_scheme}
\end{figure*}

Testing a dataset in the regression continual learning setting requires a modification of the data loading procedure during training, which can be modeled after the classification continual learning setting \cite{van2022three}. Here, data is presented to the model as a \textit{stream} rather than as the entire dataset, created by dividing the dataset into different subsets. These newly created subsets are referred to as \textit{experiences}. Three main components make up each experience: the input data, the output target values, and the task number, corresponding to the current experience, as depicted in Figure \ref{fig:CL_training_scheme}. The dataset is divided into these experiences according to a continual learning \textit{scenario}. We propose new regression CL scenarios in the Framework Overview section. 

In the CL setting, one model is incrementally trained on each experience. At each training step, the model learns from a new experience while being evaluated on all previously seen experiences, as shown in Figure \ref{fig:CL_training_scheme}. Since the model is trained incrementally, it can only access the data from the current experience, highlighting the continuous nature of this setting, where the line between the traditional 'training' and 'evaluation' phases is blurred and a more fluid scheme is adopted.

Given the scarcity of previous work on applying CL to engineering regression, our study serves two main purposes. First, as a benchmarking study, we systematically compare CL methods applied to engineering design problems, offering empirical insights into their performance. Second, we introduce continual learning to engineering regression, demonstrating its first application in mechanical design and highlighting both its potential and challenges for surrogate modeling in dynamic environments.

\section{Literature Review}

Continual Learning (CL) aims to develop models that can learn from a stream of data over time without forgetting previous knowledge, a challenge referred to as catastrophic forgetting. Much of the early and ongoing work in CL has been in the domain of classification, particularly in computer vision. Recent surveys have categorized the broad landscape of CL strategies into five major approaches: regularization-based methods that penalize changes to important parameters \cite{kirkpatrick2017overcoming}; replay-based methods that reintroduce past data during learning \cite{rolnick2019experiencereplaycontinuallearning}; architectural expansion methods; optimization-specific adjustments \cite{lopezpaz2022gradientepisodicmemorycontinual}; and representation learning approaches \cite{Wang2024CLSurvey}. These strategies have been tested extensively in well-established classification scenarios using academic benchmarks, while regression tasks and real-world engineering contexts remain underexplored.

\subsection*{Continual Learning for Traditional Classification}

Most classification CL benchmarks are organized under three canonical scenarios defined by De Lange et al. \cite{de2021continual} and further utilized by Van de Ven and Tolias: task-incremental learning, where task identities are available at test time; domain-incremental learning, where the task label is unknown but output classes remain consistent; and class-incremental learning, the most challenging, where the model must infer both the task and correct class from a shared label space without access to task identifiers. Van de Ven and Tolias used these scenarios to present a reproducible benchmark comparing major CL strategies,  using adapted versions of MNIST and CIFAR in their study  \cite{van2022three}.

Recent surveys \cite{hadsell2020embracing, mai2022online, mitchell2025continual} emphasize that the majority of progress in continual learning has been demonstrated on small-scale classification tasks, using simplified benchmarks like Split-MNIST, Permuted-MNIST, Split-CIFAR, and iCIFAR-100. These benchmarks serve as important case studies, but tend to lack the complexity of real-world data. This consideration has motivated the development of more challenging datasets and scenarios. For instance, CLEAR \cite{lin2022clear} incorporates long sequences of experiences and ambiguous task boundaries, while Stream-51 \cite{hayes2022remind} introduces a class-incremental benchmark based on high-resolution, temporal video data, bringing CL evaluation closer to real-world deployment conditions.

In general, rehearsal-based methods continue to dominate performance, with recent extensions such as MIR (Maximally Interfered Retrieval) \cite{aljundi2019online} and DER++ \cite{buzzega2020dark} improving sample efficiency and stability. However, regularization-based approaches like EWC \cite{kirkpatrick2017overcoming}, SI (Synaptic Intelligence) \cite{zenke2017continual}, and MAS (Memory Aware Synapses) \cite{aljundi2018memory} remain desirable due to their minimal memory requirements and theoretical backing.

Despite substantial work in classification CL, there are limitations to using current work in new domains. For example, methods like LwF \cite{li2017learning} and iCaRL \cite{rebuffi2017icarl} exhibit diminished performance when task boundaries are blurred or when data distributions shift continuously, as noted in recent empirical studies \cite{knoblauch2020optimal, masana2022class}.

Overall, classification remains the most thoroughly explored setting in continual learning. However, current benchmarks tend to emphasize disjoint tasks with clear boundaries, often neglecting realistic challenges such as input ambiguity, temporal effects, and complex objectives, which are prevalent in regression and real-world design problems.

\subsection*{Continual Learning Beyond Traditional Classification}
While continual learning has traditionally focused on classification with static tasks, recent work has begun exploring more complex scenarios, including regression and time-series tasks, concept drift, and different data modalities. Regression tasks introduce a distinct set of challenges for continual learning. Unlike classification, regression lacks discrete labels and may exhibit continual variation in input-output mappings over time. As Mitchell et al. \cite{mitchell2025continual} argue, the overemphasis on the classification setting limits both the theoretical development and practical deployment of CL strategies. Methods that rely on class-based memory balancing or classification-specific losses may not transfer effectively to continuous prediction problems. Verwimp et al. echoes this sentiment, introducing several open problems in continual learning, including model editing, specialization, on-device learning, and faster retraining \cite{verwimp2023continual}.
 
Some initial work has begun addressing this gap. He et al. propose the CLeaR framework for power grid forecasting under non-stationary data streams, introducing dynamic memory management and drift detection mechanisms for regression\cite{he2021clear}. Furthermore, Grote-Ramm et al. present a continual learning framework for regression networks that addresses concept drift experienced in industrial applications, combining convex optimization with memory-constrained updates \cite{grote2023continual}.

Recent work has also begun addressing continual learning in high-dimensional, structured domains, such as MIRACLE 3D \cite{resani2024miracle, resani2025continual}, which focuses on class-incremental learning for 3D point cloud data. The approach combines geometry-aware memory compression with gradient-based regularization to retain performance across tasks while significantly reducing memory usage. Though still operating in a classification setting, this method demonstrates how CL can be adapted to high-dimensional geometric inputs, which are prominent in engineering regression problems.

While these works break from traditional CL benchmarks, they remain largely focused on either classification or non-engineering regression tasks. However, many engineering problems involve regression with domain-specific constraints and data complexities, highlighting a need for more work in engineering-centered continual learning.

\subsection*{Continual Learning in Engineering Applications}
While continual learning has not yet been widely adopted in engineering design, many surrogate modeling workflows implicitly follow continual or sequential learning paradigms. Surrogate models used in optimization pipelines are frequently updated as new simulation data or experimental measurements become available \cite{palar2019use}. Examples can be found in aerodynamics (i.e., updating car drag models as electric vehicle designs evolve) or materials science (i.e., modeling metamaterials based on emerging microstructures).

In practice, however, these ML updates are often incorporated through full retraining, which Verwimp et al. \cite{verwimp2023continual} argue is wasteful, particularly when historical data vastly outweighs new data. The \textit{Machine Learning Operations (MLOps) Lifecycle}, which defines the cycle of ML development and deployment in industrial settings, involves identifying and responding to ML performance decays as incoming data characteristics change. MLOps frameworks are acknowledged as important in responding to distribution shifts; however, these shifts are largely addressed through retraining \cite{salama2021practitioners, verwimp2023continual}. Continual learning could offer an alternative approach, enabling efficient updates without discarding prior knowledge. 

Despite this conceptual alignment, systematic benchmarking of continual learning methods on real-world engineering regression problems is still lacking. Unlike classification CL, where benchmarks are standardized and deeply studied, the regression CL field has no analogous benchmark suite for evaluating general-purpose CL strategies across engineering datasets.

\subsection*{Gap and Contribution}
This work addresses current limitations in continual learning for engineering by assembling a benchmark suite for continual learning in engineering regression contexts. We evaluate several established CL strategies, including Experience Replay (ER), Elastic Weight Consolidation (EWC), and Gradient Episodic Memory (GEM), on regression tasks involving high-dimensional, geometric data for engineering design. Our benchmark targets the challenges of continual surrogate learning, such as distributional drift, high-dimensionality, and multimodal input representations (i.e., point clouds and parametric descriptions). Unlike prior regression CL studies that use synthetic data or focus on single-stream updates, our benchmarks draw from realistic, diverse datasets and provide a foundation for testing general-purpose CL methods in engineering domains. In doing so, we help bridge the gap between theoretical advances in CL and practical needs of engineering design systems that operate under dynamic, evolving data regimes.


\section{Framework Overview}

To properly position the regression, engineering domain in the continual learning setting, we introduce our selected engineering datasets, corresponding surrogate models, and the new regression CL scenarios in which we will benchmark the chosen strategies.

\subsubsection*{Datasets}

We compile five datasets spanning naval architecture, automotive aerodynamics, and aircraft aerodynamics. Examples of geometries from each dataset can be seen in Figure \ref{fig:CL_training_scheme}. In each dataset, the data are represented as point clouds, a common data representation for engineering problems that captures the geometry of a design by plotting the shape as 3D points in space. Two of the selected datasets, namely the SHIPD and DrivAerNet datasets, also offer parametric representations of their data, where the geometry is defined by a vector of design parameters; thus, these datasets are tested using both their point cloud and parametric data representations. See Appendix A for more detailed information on each dataset. 

The SHIPD dataset \cite{bagazinski2023ship, DVN/MMGAUS_2024} contains 10,000 geometric ship hull designs and their associated wave drag coefficients. We use both the point cloud and parametric representations in our study. 
The DrivAerNet \cite{elrefaie2025drivaernet} and DrivAerNet++ \cite{elrefaie2025drivaernet++} datasets contain car geometries and their associated drag values. DrivAerNet includes 4,000 car designs in the Fastback car category as point clouds. DrivAerNet++ includes 4,000 additional cars, with designs from the Fastback, Notchback, and Estateback car categories, offered as both parametric data and point clouds.
The ShapeNet Car dataset \cite{song2023surrogate} contains 9,070 car designs as point clouds, along with their associated drag coefficients.
The Rapid Aerodynamic and Deep Learning (RAADL) dataset, provided by MIT Lincoln Laboratory \cite{jones2024rapid}, consists of 800 glider geometries represented as point clouds, along with their associated drag coefficients.

 \subsubsection*{Continual Learning Scenarios in Regression}
The continual learning setting requires models to incrementally learn from a stream of data, without access to previously seen data. In the academic continual learning setting proposed in \cite{de2021continual}, the methods of splitting datasets into experiences are referred to as \textit{scenarios}. Common scenario breakdowns for classification include dividing datasets based on classes (class-incremental) or tasks, such as distinguishing between two different classes (task-incremental). Because regression does not have explicit class distinctions, new scenarios must be defined. Here, we propose three CL scenarios --- \textit{bin incremental}, \textit{input incremental}, and \textit{multi-target incremental} --- that adapt the classification scenarios to the engineering regression setting.

\begin{figure*}[t]
\centering
\begin{subfigure}[t]{0.3\textwidth}
    \centering
    \includegraphics[width=\linewidth]{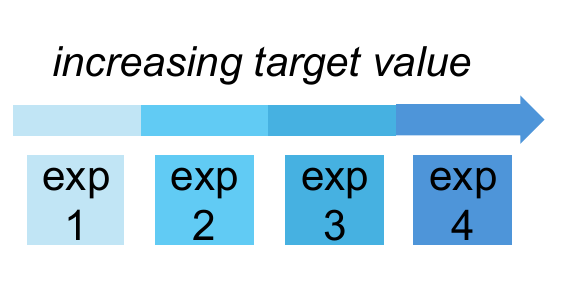}
    \caption{Bin Incremental}
    \label{fig:bin_inc}
\end{subfigure}
\hfill
\begin{subfigure}[t]{0.3\textwidth}
    \centering
    \includegraphics[width=\linewidth]{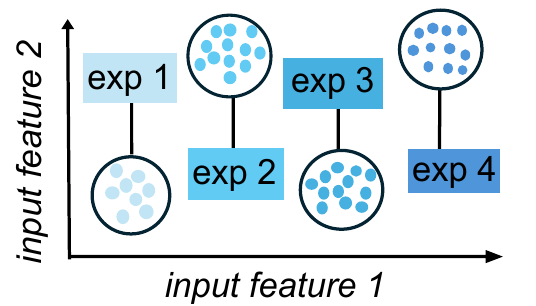}
    \caption{Input Incremental}
    \label{fig:input_inc}
\end{subfigure}
\hfill
\begin{subfigure}[t]{0.3\textwidth}
    \centering
    \includegraphics[width=\linewidth]{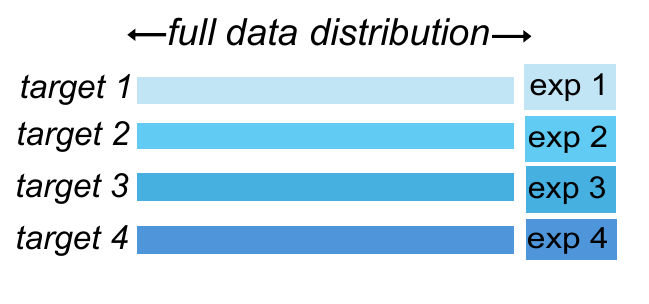}
    \caption{Multi-target Incremental}
    \label{fig:multi_targ}
\end{subfigure}
\label{fig:describe_cl_scenarios}
\caption{Proposed regression continual learning scenarios that would be relevant for engineering. Figure \ref{fig:bin_inc} depicts the bin incremental scenario, in which the data is divided into experiences based on binning the target values. Figure \ref{fig:input_inc} depicts the input incremental scenario, in which the data is divided into experiences based on input categorizations, shown as clusters of the input data in the figure. Figure \ref{fig:multi_targ}  depicts the multi-target incremental scenario, in which the data is divided into experiences based on newly desired targets, meaning each experience contains a full dataset, but a new predictive target.}
\end{figure*}

\paragraph{Bin Incremental}
In this scenario, the continuous output space is partitioned into ranges or “bins,” and the model learns to predict outputs in one range at a time (analogous to class-incremental learning, but for continuous values). For example, a model predicting drag coefficients might first be trained on cases with low drag values, then incrementally extended to moderate and high drag cases. Each bin introduces a new subset of the regression target space. This setup is useful when different regimes of the output might require different model behaviors (e.g., low vs. high Reynolds number flow regimes in a fluid dynamics surrogate). It tests whether a model can extend its prediction range without forgetting how to predict in previously learned ranges. Bin-incremental setups are valuable for benchmarking because they create distinct ``tasks'' out of a continuous variable, forcing the model to handle incremental complexity. They also simulate an engineer progressively refining a model’s valid operating range (for instance, gradually increasing the load conditions that a structural surrogate can handle). For the benchmarks we test in this scenario, the target values are divided into \textit{n} bins, where \textit{n} is the number of desired experiences. Each bin is then assigned to a different experience, and the model is incrementally trained accordingly. This paper's investigation evaluates all datasets in this scenario. Figure \ref{fig:bin_inc} visually depicts this scenario.

\paragraph{Input Incremental}
Here, the input distribution or the underlying data generation process changes over time, and the model must continually adapt. Unlike bin-incremental, the target quantity may remain the same, but the relationship between input and output shifts. Formally, ``the underlying data generation process is changing over time due to non-stationarity of the data stream'', meaning either the input distribution $P(X)$ changes or the output function $P(Y|X)$ changes (or both) \cite{mitchell2025continual}. In engineering terms, this could correspond to a concept drift in sensor data or operating conditions. For example, consider a predictive maintenance model for a turbomachine: initially it sees data from a brand-new machine, but over time, the machine ages, and sensor readings (inputs) for the same health indicators (output) shift distribution. The model should update to this new regime. Another example is climate effects on wind turbine performance – the input wind patterns shift gradually due to climate change, altering the power output relationship. In our benchmarks, input incremental scenarios are created by splitting data by operating conditions or input category. The model is trained sequentially on each segment. Crucially, there is no clear task boundary in terms of output type – the tasks are distinguished by the domain of inputs. The challenge for CL algorithms is to adjust to the new domain without degrading performance on the old domain. 
In our experiments, categorizations of the input data are used to define different experiences. We test the DrivAerNet++ dataset in this scenario. Figure \ref{fig:input_inc} visually depicts this scenario.

\paragraph{Multi-target Incremental}
In this scenario, the model faces new output targets over time , effectively learning additional regression tasks that may share the same input space. This reflects cases where new metrics or phenomena need to be predicted as a project progresses. Initially, a model might predict one quantity from input data; later, a new quantity is introduced and the model must learn to predict that as well, alongside the original target. Engineering workflows where this is relevant include iterative simulation campaigns that add new quantities of interest, and evolving design criteria (adding new objectives or constraints that require predictive models). This study does not benchmark in the multi-target incremental scenario as we do not include model architecture strategies in our investigation; however, we set this up as a necessary scenario for engineering CL to be explored in future work. Figure \ref{fig:multi_targ} visually depicts this scenario.

\subsubsection*{Benchmarks}
\begin{figure*}[h]
  \centering
  \begin{subfigure}{\textwidth}
    \includegraphics[width=\linewidth]{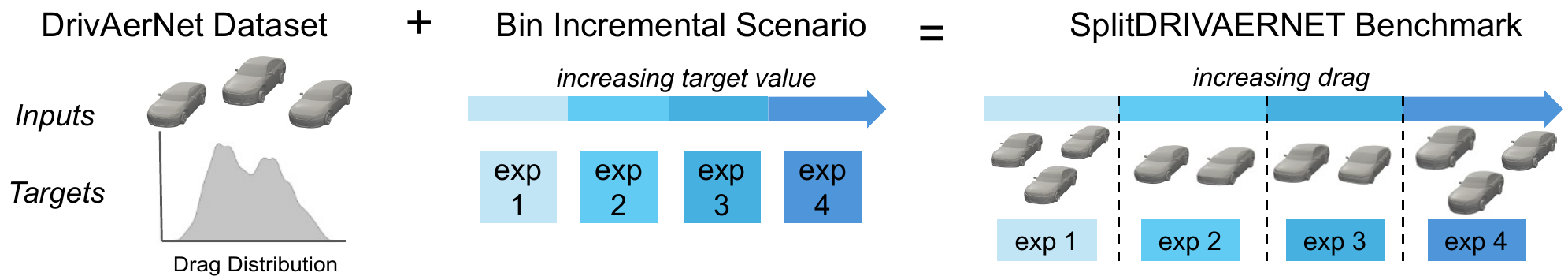}
    \caption{The SplitDRIVAERNET benchmark is created by combining the DrivAerNet dataset with the bin incremental scenario. In this case, each experience consists of car geometry samples with similar drag coefficients.}
    \label{fig:bin_inc_example}
  \end{subfigure}
  \hfill
  \begin{subfigure}{\textwidth}
    \includegraphics[width=\linewidth]{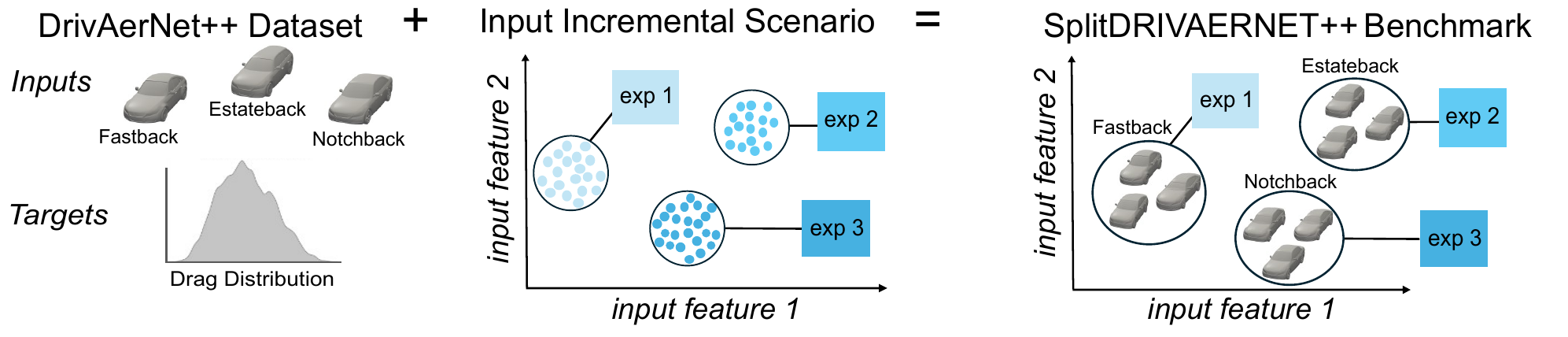}
    \caption{The SplitDRIVAERNET++ benchmark is created by combining the DrivAerNet++ dataset with the input incremental scenario. In this case, each experience consists of car geometry samples from each one of the following car categories: Fastback, Estateback, and Notchback.}
    \label{fig:input_inc_example}
  \end{subfigure}
  \caption{Continual learning benchmarks are defined as the combination of a dataset with a continual learning scenario. }
  \label{fig:benchmark_explanation}
\end{figure*}

The provided benchmarks have been generated by combining the engineering datasets described above with one of the regression CL scenarios. To create these benchmarks, a dataset is divided into experiences based on the chosen CL scenario and each data sample is assigned a 'task number' that corresponds to its respective experience (see Figure \ref{fig:CL_training_scheme}); this ensures that the model is trained with the appropriate data from each experience stream. Examples of benchmarks created using the bin incremental scenario (illustrated with the DrivAerNet dataset) and the input incremental scenario (illustrted with the DrivAerNet++ dataset) can be seen in Figure \ref{fig:benchmark_explanation}. Table \ref{tab:benchmark_creation_info} outlines all created benchmarks and relevant information, including the dataset, data representation type, model, and CL scenario.

\begin{table*}[htbp]
\centering
\small
\setlength{\tabcolsep}{3pt}
\renewcommand{\arraystretch}{1}
\begin{tabular}{l l l l l l}
\toprule
        \textbf{Benchmark} & \textbf{Dataset} & \textbf{Representation} & \textbf{Model} & \textbf{CL Scenario} \\
        \midrule
        SplitSHIPD-Par & SHIPD & Parametric & ResNet & Bin Incremental \\
        SplitSHIPD-PC & SHIPD & Point clouds & Pretrained RegPointNet & Bin Incremental \\
        SplitSHAPENET & ShapeNet Car Drag & Point clouds & Pretrained RegPointNet & Bin Incremental \\
        SplitRAADL & RAADL & Point clouds & Pretrained RegPointNet & Bin Incremental \\
        SplitDRIVAERNET-PC & DrivAerNet & Point clouds & Pretrained RegPointNet & Bin Incremental \\
        SplitDRIVAERNET++-Par & DrivAerNet++ & Parametric & ResNet & Bin Incremental and Input Incremental \\
        SplitDRIVAERNET++-PC & DrivAerNet++ & Point clouds & Pretrained RegPointNet & Bin Incremental and Input Incremental \\
        \bottomrule
    \end{tabular}
    \caption{Provided benchmarks and their associated datasets, representations, model, and CL scenarios.}
    \label{tab:benchmark_creation_info}
\end{table*}

We created seven benchmarks using the bin incremental scenario and two using the input incremental scenario, selected based on how each dataset was structured. The DrivAerNet++ dataset  has explicit input boundaries based on the different types of cars (Fastback, Estateback, and Notchback), and therefore, its point cloud and parametric representations were tested in the input incremental scenario as well as the bin incremental scenario.

\section{Methodology}

 \subsubsection*{Surrogate Models} 
 Two types of surrogate model architectures are used in our experiments to handle the different data representations. A regression PointNet architecture is used for all point cloud representation benchmarks and a ResNet50 architecture is used for both parametric representation benchmarks.  The hyperparameter tuning of each surrogate model was conducted using a hit-and-trial approach, adjusting model-specific hyperparameters such as learning rate, number of epochs, loss function, batch size, and model architecture. Models were evaluated on a separate validation set outside the continual learning setting to optimize their baseline performance for benchmarking. 

 \paragraph{Regression PointNet} A regression PointNet model architecture is trained on all point cloud data benchmarks. PointNet is a well-established architecture designed to process unordered point clouds \cite{qi2017pointnet}. We adapt PointNet for regression by modifying its output layer to predict continuous values.To address over-fitting issues due to data constraints, we use a pretrained PointNet model from Model Zoo \cite{PointNeXtModelZoo} that was trained on the ScanObjectNN dataset, which consists of point cloud representations of scanned indoor scene data. The PointNet architecture consists of Spatial Transformer Networks (STN), an encoder, and a regression head. We freeze several internal model layers and only retrain the outer parameters. This reduction in trainable parameters allows the model to learn from the smaller amount of data without over-fitting and provides the model with a strong weight initialization that is already tuned to point cloud data.
 
\paragraph{Regression ResNet} The SplitSHIPD-Par and SplitDRIVAERNET++-Par benchmarks are trained using a regression ResNet model \cite{he2016deep}. ResNet is a deep convolutional neural network architecture that introduced residual connections to address vanishing gradients in deep networks. These connections allow information to bypass layers through identity mappings when training deep networks. For our regression tasks, we modified the standard classification architecture by replacing the final softmax layer with a linear output layer to predict continuous values, while retaining the feature extraction capabilities of the original network.

\subsubsection*{Continual Learning Strategies}
\begin{figure*}[t]
\centering
\begin{subfigure}[t]{0.25\textwidth}
    \centering
      \includegraphics[width=\textwidth]{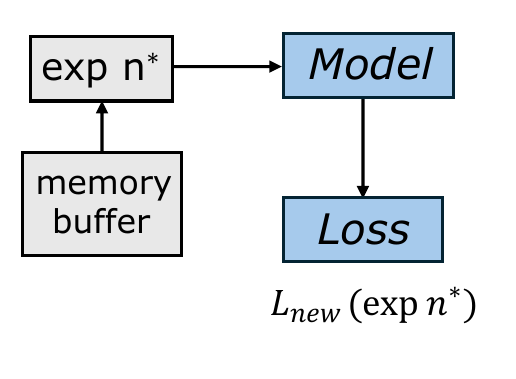}
      \caption{Experience Replay}
      \label{fig:replay_fig}
\end{subfigure}
\hfill
\begin{subfigure}[t]{0.3\textwidth}
    \centering
  \includegraphics[width=\textwidth]{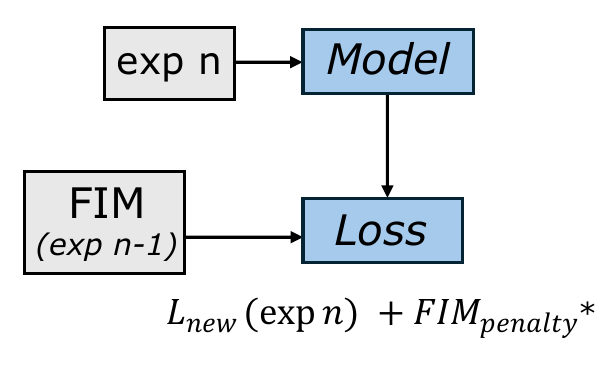}
  \caption{Elastic Weight Consolidation}
  \label{fig:ewc_fig}
\end{subfigure}
\hfill
\begin{subfigure}[t]{0.33\textwidth}
    \centering
      \includegraphics[width=\textwidth]{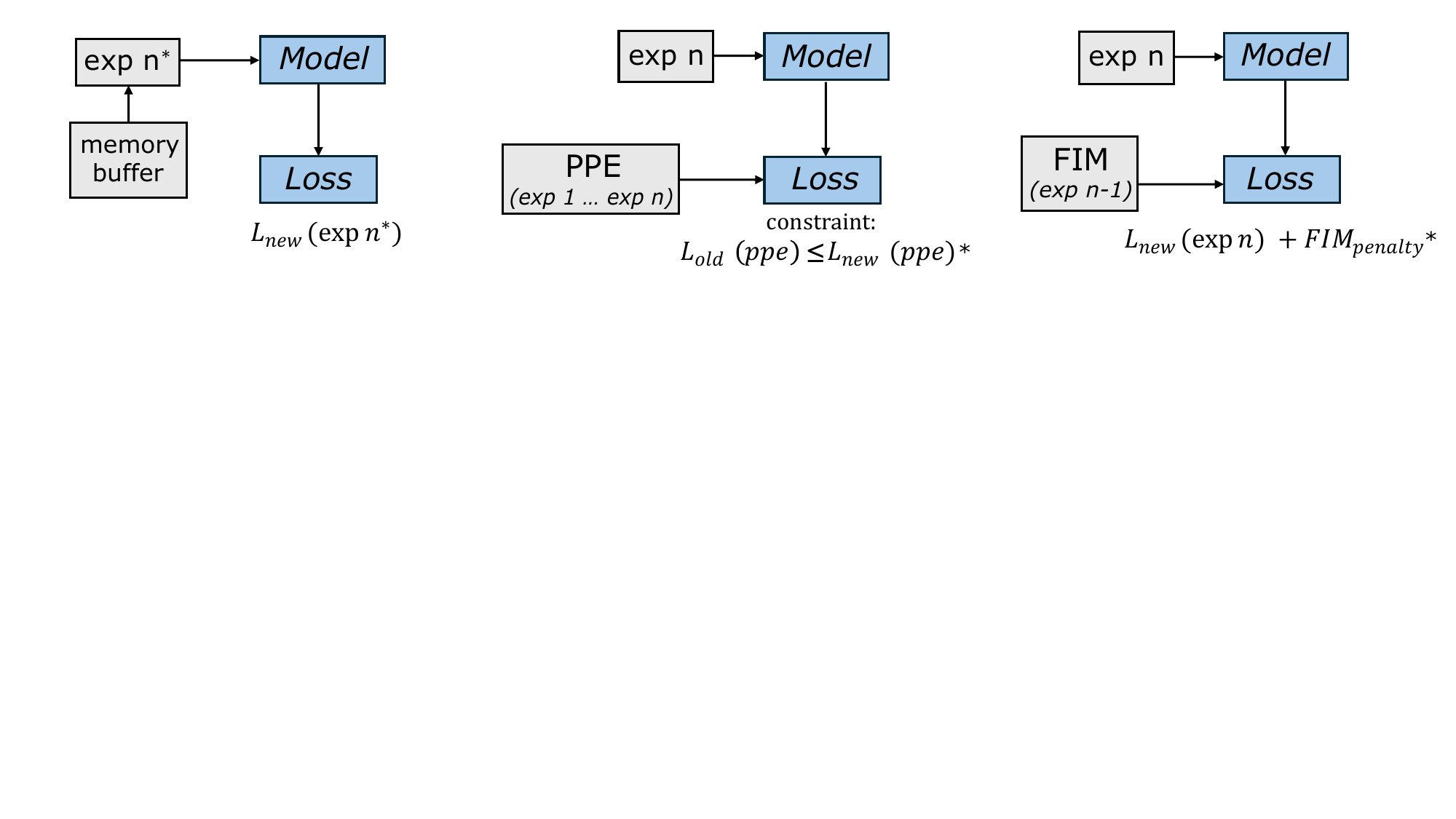}
      \caption{Gradient Episodic Memory}
      \label{fig:gem_fig}
\end{subfigure}
\caption{Overview of the continual learning strategies evaluated in this work. In Figure \ref{fig:replay_fig}, Experience Replay (ER) reintroduces stored samples from previous experiences during training via a memory buffer. In Figure \ref{fig:ewc_fig}, Elastic Weight Consolidation (EWC) adds a regularization term based on the Fisher Information Matrix to preserve important parameters from earlier tasks. In Figure \ref{fig:gem_fig}, Gradient Episodic Memory (GEM) constrains the loss function using stored samples from previous experiences, based on a specified number of patterns per experience (PPE). The * symbol indicates where past knowledge is integrated into the training process for each method.}
\label{fig:cl_strategies}
\end{figure*}

Several continual learning strategies have been developed in the classification setting with the primary objective of mitigating catastrophic forgetting without retraining on all previously seen data. Existing strategies have been categorized in several ways, with many studies identifying overarching strategy groups of rehearsal-based and regularization-based methods \cite{de2021continual, Wang2024CLSurvey, van2022three}. The three strategies tested in this study include Experience Replay (ER), Elastic Weight Consolidation (EWC), and Gradient Episodic Memory (GEM).

\paragraph{Experience Replay (ER)} ER is a foundational CL strategy that takes a data-driven approach to the catastrophic forgetting problem. As seen in Figure \ref{fig:replay_fig}, this method stores a subset of past data in a memory buffer and mixes it with new data during training to preserve knowledge of previous tasks. In practice, a randomly selected group of samples from each experience is retained in memory, which is updated at the end of each experience and replayed at the beginning of all subsequent ones \cite{rolnick2019experiencereplaycontinuallearning}. ER is simple and computationally efficient, requiring minimal additional compute overhead. However, it incurs a moderate memory cost that scales with the size of the buffer.

\paragraph{Elastic Weight Consolidation (EWC)} EWC is a regularization-based approach to mitigating catastrophic forgetting in neural networks \cite{kirkpatrick2017overcoming}. Unlike ER and GEM, EWC does not store old data, but instead tracks parameter importance using the diagonal of the Fisher Information Matrix (FIM). As seen in Figure \ref{fig:ewc_fig}, it penalizes weight updates that would significantly affect parameters deemed important to previous tasks. The regularized loss becomes \( L_{\text{EWC}} = L_{\text{new}} + \frac{\lambda}{2} \sum_i F_i (\theta_i - \theta_i^*)^2 \), where \( L_{\text{new}} \) is the loss on the current task, \( \theta_i^* \) are the previously learned parameters, \( F_i \) is the Fisher importance, and \( \lambda \) is a hyperparameter that controls regularization strength. EWC has a low memory footprint but adds computational overhead that comes from estimating and storing importance weights.

\paragraph{Gradient Episodic Memory (GEM)} GEM is an optimization-based strategy, often categorized under rehearsal-based constraint methods, that aims to directly control forgetting by operating in gradient space \cite{lopezpaz2022gradientepisodicmemorycontinual}. It stores past data in an episodic memory like ER but differs in how it uses this data. Instead of simply replaying it, GEM uses a subset of past data, referred to as the patterns per experience (PPE), to constrain the loss function by ensuring gradient updates for current data do not interfere with model performance on previous data, represented by the PPE (see Figure \ref{fig:gem_fig}. Specifically, it solves a quadratic program that aligns the current gradient with those from memory, preserving or improving past performance. GEM is more compute-intensive than ER or EWC, due to gradient projection and per-task constraint checks, and has a higher memory footprint, as it requires storage of both data and associated gradients for each past task.

\subsection*{Implementation and Evaluation Metrics}

All experiments are conducted using the Pytorch Continual Learning library called Avalanche \cite{lomonaco2021avalanche}, which provides a framework for benchmarking classification CL strategies. The library provides existing scenario implementations as well as templates for implementing a new scenario. It also offers several continual learning strategies, including the strategies of Replay, EWC, and GEM tested in this study, that can be evaluated on the created benchmarks. Logging and evaluation tools are provided for tracking selected metrics for each experiment.

To constrain and evaluate the benchmarks in the continual learning setting, we select metrics relevant to evaluating regression models and tracking the retention of prior tasks. To appropriately reflect the CL setting, we constrain our problems based on computational resources and time. All experiments are allowed the same amount of compute, offered by two Quadro RTX 5000 GPUs and an Intel Xeon Silver 4215R CPU (8 cores, 16 threads, 3.20 GHz base, 4.00 GHz max) with 62.43 GB RAM. Furthermore, we constrain rehearsal strategies with a data storage limitation of 20\% of the dataset and enforce that all methods have a shorter total runtime than the time required to retrain on all data (denoted as the Joint strategy in results). Traditional continual learning benchmarks are evaluated on accuracy and forgetting. We adapt these metrics for the regression setting, but aim to measure the same capabilities as its classification analog. High performing strategies are marked by a reduction in forgetting between the Naive baseline and the tested strategy as well as a low prediction error on the entire test set and each individual experience. These metrics are further motivated and described below.

\paragraph{Error Metrics (MPE and MAE)} In Tables~\ref{tab:benchmark_comparison} and \ref{tab:benchmark_comparison_input}, we report the Mean Percent Error (MPE) to enable fair comparisons across datasets with different scales. MPE is computed by dividing the absolute error by the true value for each sample, averaging across all samples, and multiplying by $100$ to express it as a percentage.

We also report the Mean Absolute Error (MAE), which computes the average error between predictions $\hat{y}$ and targets $y$, defined as $\text{MAE} = \frac{1}{n} \sum_{i=1}^{n} |y_i - \hat{y}_i|$, where $n$ is the number of samples. MAE provides a direct, interpretable measure of error in the same units as the target variable, making it especially useful in engineering contexts.

In our benchmark-specific tables (Tables~\ref{tab:comparison_splitshipPar}-\ref{tab:comparison_bin_splitdrivaernet++PC} and Tables~\ref{tab:comparison_splitdrivaernet++Par} and \ref{tab:comparison_splitdrivaernet++PC}), we report raw MAE values for predictions of the final model on each experience for each strategy. To track incremental performance as new experiences are added, depicted in Figure \ref{fig:all-benchmarks-grid}, we compute the incremental MAE after each training stage. Specifically, after learning up to experience $j$, we evaluate the model on all experiences seen so far and compute the average MAE across these experiences: $\overline{\text{MAE}}_{j} = \frac{1}{j} \sum_{k=1}^{j} \left(\frac{1}{n_k} \sum_{i=1}^{n_k} \left| y_{k,i} - \hat{y}_{j,k,i} \right|\right)$ where $n_k$ is the number of samples in experience $k$, $y_{k,i}$ is the true value for sample $i$ in experience $k$, and $\hat{y}_{k,i}$ is the predicted value for sample $i$ in experience $k$. This metric captures both the model's retention of knowledge from previous experiences and its ability to learn new tasks, providing a comprehensive view of continual learning performance over time.

\paragraph{Forgetting} Forgetting quantifies how much performance degrades on previously learned experiences as new ones are introduced. In continual learning, this metric captures the extent of catastrophic forgetting, a key issue in engineering contexts where past knowledge remains critical. For classification, forgetting is defined by how much accuracy decreases on previous tasks once a model is trained on new tasks. For regression, we adapt this by replacing accuracy with mean absolute error (MAE), defining forgetting on an experience $j$ after training on experience $k$ as:
\[
f_{k,j} = \max_{l \in \{j, \dots, k-1\}} \left| \text{MAE}_{l,j} - \text{MAE}_{k,j} \right| \quad \text{for all } j < k,
\]
where $j$ is the experience being evaluated, $k$ is the current training stage, and $l$ refers to all training stages between experience $j$ and $k$ (i.e., $j \leq l < k$). Thus, $\text{MAE}_{l,j}$ is the mean absolute error on experience $j$ after training on experience $l$ and $\text{MAE}_{k,j}$ is the error on the same experience $j$ after training through current experience $k$. Iterating over all experiences $l$, this metric measures the maximum performance degradation on a given experience $j$ after encountering new experiences.

To more easily compare values across different error scales (different datasets and experiences), we normalize forgetting to arrive at the forgetting ratio (FR), defined as $FR_{k,j} = f_{k,j} / \text{MAE}_{j,j}$, where $\text{MAE}_{j,j}$ is the model's error on experience $j$ when it was first learned. We use the forgetting ratio in all benchmark-specific tables.

To summarize forgetting across all experiences, we compute the average forgetting ratio $\overline{FR}_k = \frac{1}{k-1} \sum_{j=1}^{k-1} FR_{k,j}$, and use the lowest $\overline{FR}_k$ achieved over all trials as the best forgetting score per strategy in Tables ~\ref{tab:benchmark_comparison} and \ref{tab:benchmark_comparison_input}.

For visualizations, we use the average absolute forgetting $\overline{f}_k = \frac{1}{k-1} \sum_{j=1}^{k-1} f_{k,j}$ on experiences $j$ through $k-1$ after each experience $k$ as our incremental forgetting measure (see Figure~\ref{fig:all-benchmarks-grid}), which intuitively depicts the raw increase in error on past tasks as training progresses.

\paragraph{Train time} We report the average total run times required to complete a full continual learning experiment for each benchmark, which includes the continuous training and evaluation phases across the entire stream of experiences (see Table~\ref{tab:timing}). Continual learning strategies differ in computational complexity. Regularization-based methods, such as EWC, introduce minimal memory overhead but incur additional compute for estimating parameter importance. In contrast, rehearsal-based methods require time to replay stored samples during training. To assess the practical efficiency of continual learning strategies, we compare their total run time against the Joint retraining baseline. This highlights the potential computational savings of continual learning approaches relative to retraining from scratch.

\section{Results}
We evaluated three continual learning strategies (Replay, GEM, EWC) across various engineering datasets, input modalities (parametric and point cloud), and continual learning scenarios (bin incremental and input incremental). For each benchmark, the three strategies are compared based on each strategy's final MPE and average forgetting ratio for the full test set (Tables~\ref{tab:benchmark_comparison} and \ref{tab:benchmark_comparison_input}), their MAEs and forgetting ratios for individual experiences (Tables~\ref{tab:comparison_splitshipPar}-\ref{tab:comparison_bin_splitdrivaernet++PC} and Tables~\ref{tab:comparison_splitdrivaernet++Par} and \ref{tab:comparison_splitdrivaernet++PC}), and their incremental MAE and forgetting on experiences it encountered within the stream (Figure~\ref{fig:all-benchmarks-grid}). Results are presented first for the bin incremental scenario, and subsequently for the input incremental scenario, allowing clear comparisons across datasets and methods.

\subsection*{Bin Incremental Benchmark Outcomes} 
Table \ref{tab:benchmark_comparison} compares all strategies based on MPE and average forgetting ratio across all datasets tested in the bin incremental scenario, which are discussed as overall trends. Then, results for each benchmark are detailed, including each strategy's incremental MAE and forgetting, as well as the final model's performance on each experience. Computational efficiency is also discussed.

\subsubsection*{Overall Trends}
Across all datasets, Replay consistently achieved the best performance among continual learning strategies, demonstrating lower MPE than GEM, EWC, and Naive approaches. On average, Replay closely approached the cumulative baseline, notably evident in SplitDRIVAERNET (2.07\% vs. 1.57\%) and SplitDRIVAERNET++-PC (3.64\% vs. 3.36\%). GEM provided intermediate performance, improving upon EWC and the Naive method but generally lagging behind Replay. It notably exhibited low forgetting ratios in some benchmarks, such as SplitSHIPD-Par (0.24), SplitSHAPENET (0.39), and SplitRAADL (-0.49), suggesting selective effectiveness in mitigating forgetting. EWC consistently underperformed across all datasets, closely matching the Naive scenario, underscoring its limited effectiveness.

Regarding computational efficiency (see Table~\ref{tab:timing}), EWC was often the fastest strategy due to its minimal computational overhead. In contrast, GEM was generally the slowest strategy, although it was notably faster on the small RAADL dataset and the lower-dimensional parametric datasets compared to its typical performance. Replay typically offered fast computational times, outperforming EWC on two benchmarks, but experienced slower performance on larger, high-dimensional point cloud benchmarks. Despite these exceptions, Replay consistently demonstrated the strongest balance between accuracy and computational efficiency.

The following sections provide detailed results for each benchmark, supporting these trends and discussing any notable deviations.

\begin{table*}[h]
    \centering
        \small
    \renewcommand{\arraystretch}{0.9}
    \setlength{\aboverulesep}{0.5ex}
    \setlength{\belowrulesep}{0.5ex}
    \begin{tabular}{l *{5}{c} @{\hspace{2em}} *{4}{c}}
        \toprule
        \multicolumn{1}{c}{\textbf{Metrics}} & \multicolumn{5}{c}{\textbf{Mean Percent Error} ↓} & \multicolumn{4}{c}{\textbf{Forgetting Ratio} ↓}\\
        \midrule
        \multicolumn{1}{c}{\textbf{Strategy}} & \textbf{Baseline} & \textbf{Replay} & \textbf{GEM} & \textbf{EWC} & \textbf{Naive} & \textbf{Replay} & \textbf{GEM} & \textbf{EWC} & \textbf{Naive}\\
        \midrule
        SplitSHIPD-Par & $8.11$ & $\mathbf{11.51}$ & $15.15$ & $25.16$ & $25.66$ & $0.56$ & $\mathbf{0.24}$ & $4.83$ & $4.94$ \\
        SplitSHIPD-PC & $11.21$ & $\mathbf{12.75}$ & $17.66$ & $23.22$ & $25.44$ & $\mathbf{0.40}$ & $0.47$ & $3.53$ & $3.79$ \\
        \midrule
        SplitSHAPENET & $6.16$ & $\mathbf{7.62}$ & $13.57$ & $23.15$ & $24.25$ & $0.89$ & $\mathbf{0.39}$ & $7.95$ & $8.89$ \\
        \midrule
        SplitRAADL & $4.98$ & $\mathbf{6.95}$ & $7.75$ & $11.96$ & $57.81$ & $\mathbf{-0.57}$ & $-0.49$ & $-0.39$ & $1.53$ \\
        \hline
        SplitDRIVAERNET & $1.57$ & $\mathbf{2.07}$ & $6.08$ & $9.24$ & $9.33$ & $\mathbf{-0.01}$ & $0.66$ & $3.63$ & $4.90$
        \\
        \midrule
        SplitDRIVAERNET++-Par & $3.13$ & $\mathbf{4.90}$ & $7.20$ & $10.18$ & $10.57$ & $\mathbf{0.51}$ & $0.98$ & $3.01$ & $3.38$ \\
        SplitDRIVAERNET++-PC & $3.36$ & $\mathbf{3.64}$ & $6.97$ & $10.66$ & $13.22$ & $\mathbf{0.35}$ & $0.37$ & $ 5.16 $ & $7.23$ \\
        \bottomrule
    \end{tabular}
    \caption{Comparison of final Mean Percent Error and Best Forgetting Ratio across different benchmarks and CL methods for the bin incremental scenario}
    \label{tab:benchmark_comparison}
\end{table*}

\begin{figure*}[!tb]
  \centering

  \begin{subfigure}[b]{0.19\textwidth}
    \includegraphics[width=\textwidth]{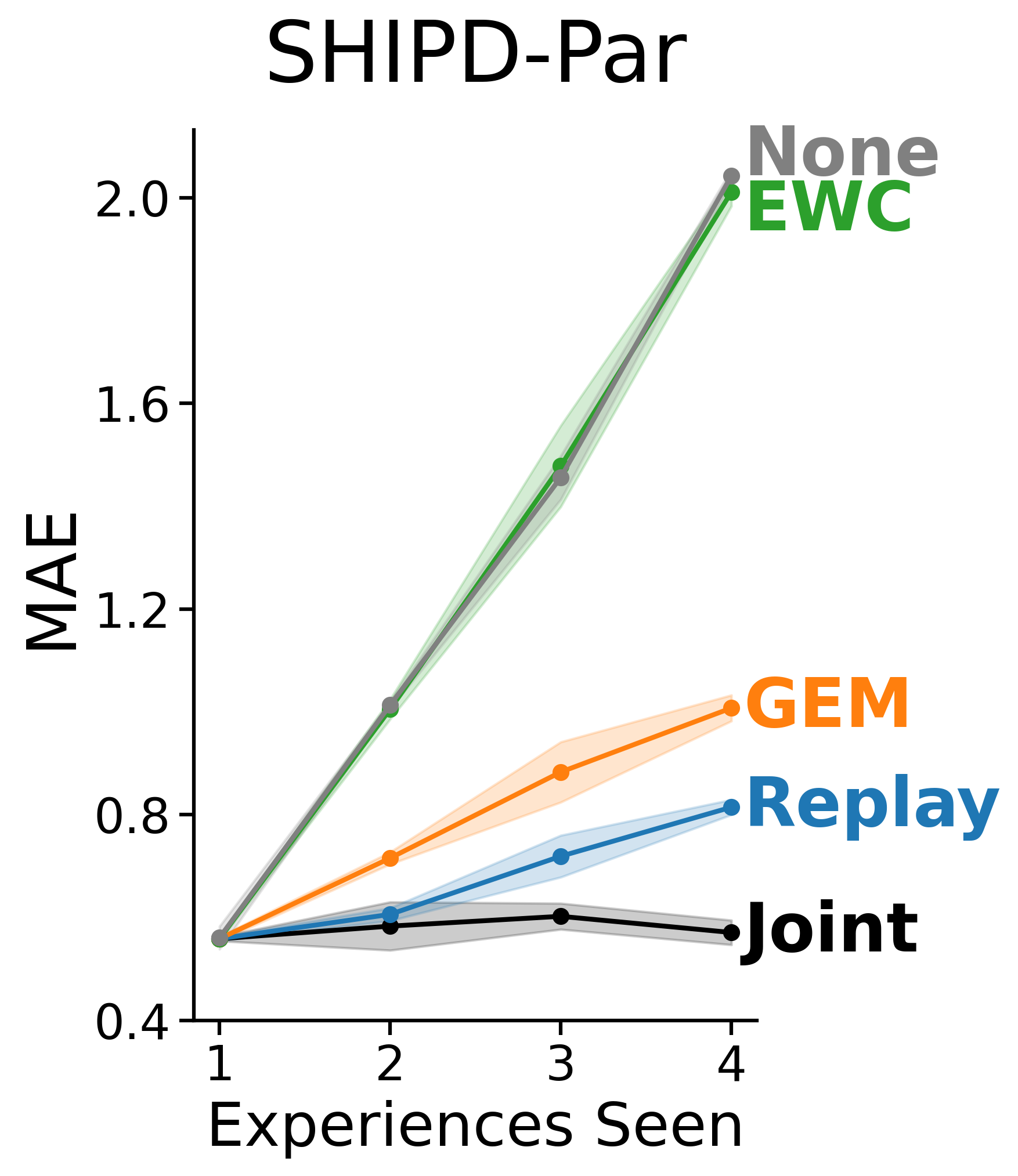}
    \caption{}
    \label{fig:SHIPDsub-Par_mae}
  \end{subfigure}
  \begin{subfigure}[b]{0.19\textwidth}
    \includegraphics[width=\textwidth]{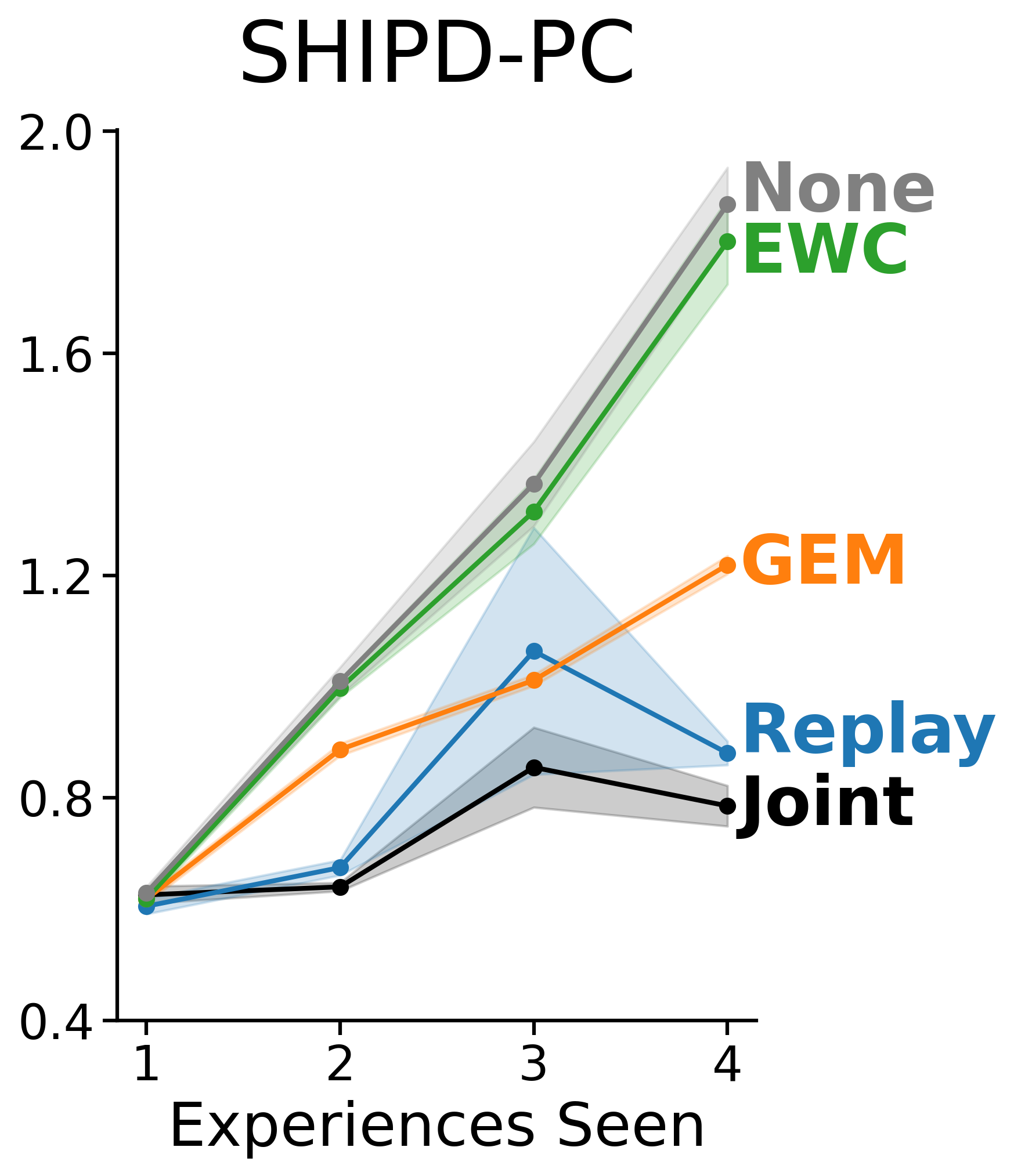}
    \caption{}
    \label{fig:SHIPD-PC_mae}
  \end{subfigure}
  \begin{subfigure}[b]{0.19\textwidth}
    \includegraphics[width=\textwidth]{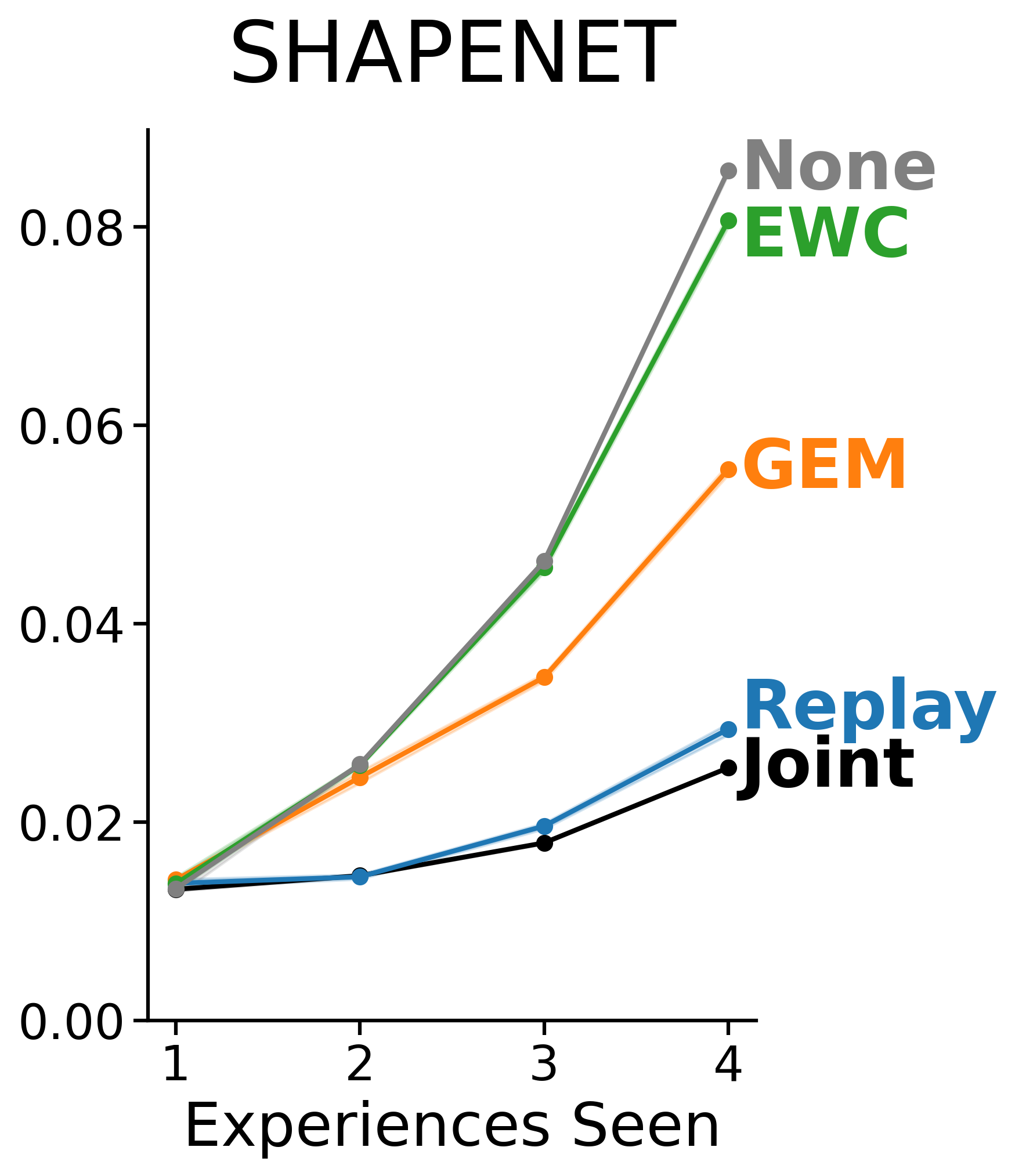}
    \caption{}
    \label{fig:SHAPENET-PC_mae}
  \end{subfigure}
  \begin{subfigure}[b]{0.19\textwidth}
    \includegraphics[width=\textwidth]{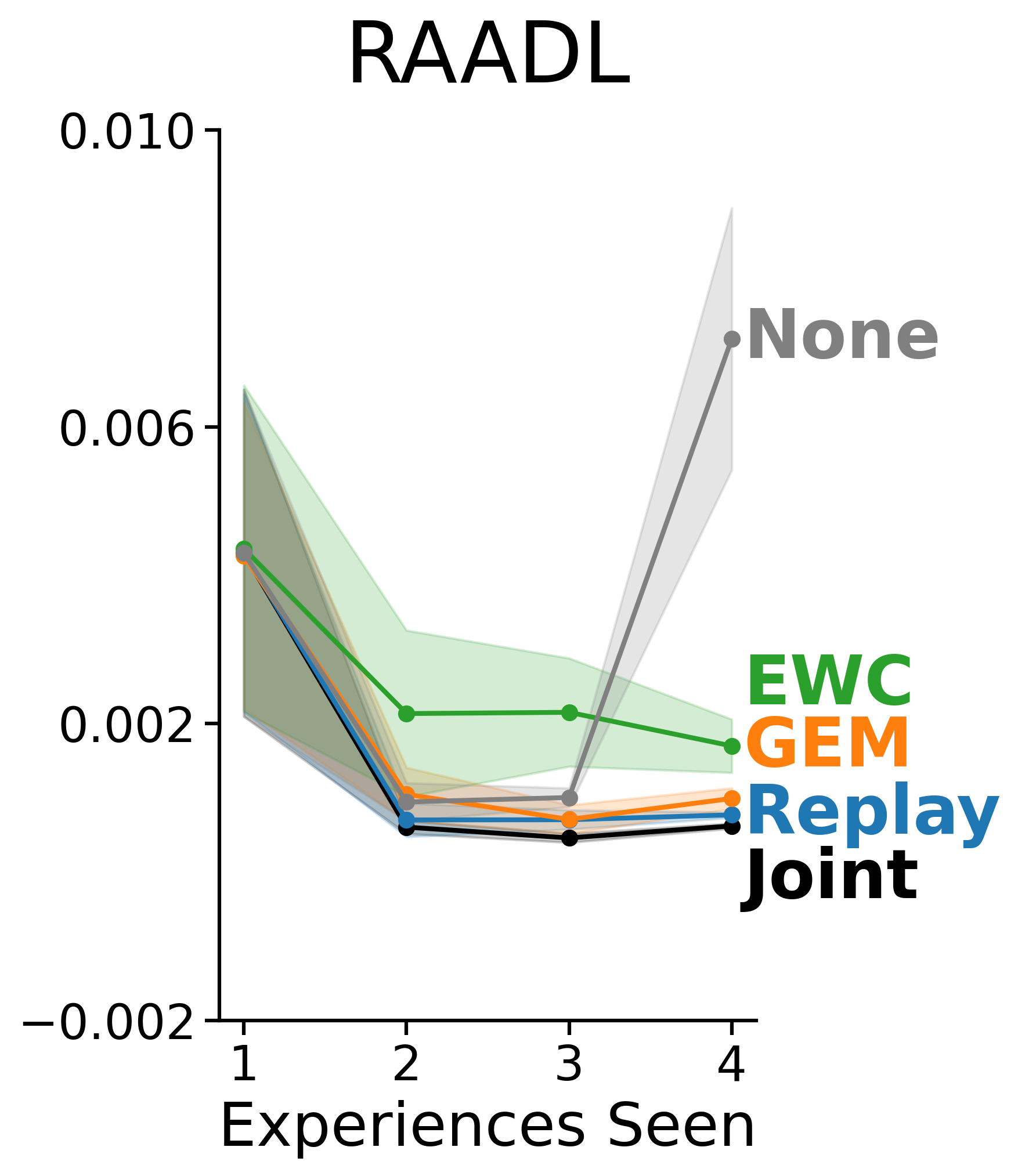}
    \caption{}
    \label{fig:RAADL-PC_mae}
  \end{subfigure}
  \begin{subfigure}[b]{0.19\textwidth}
    \includegraphics[width=\textwidth]{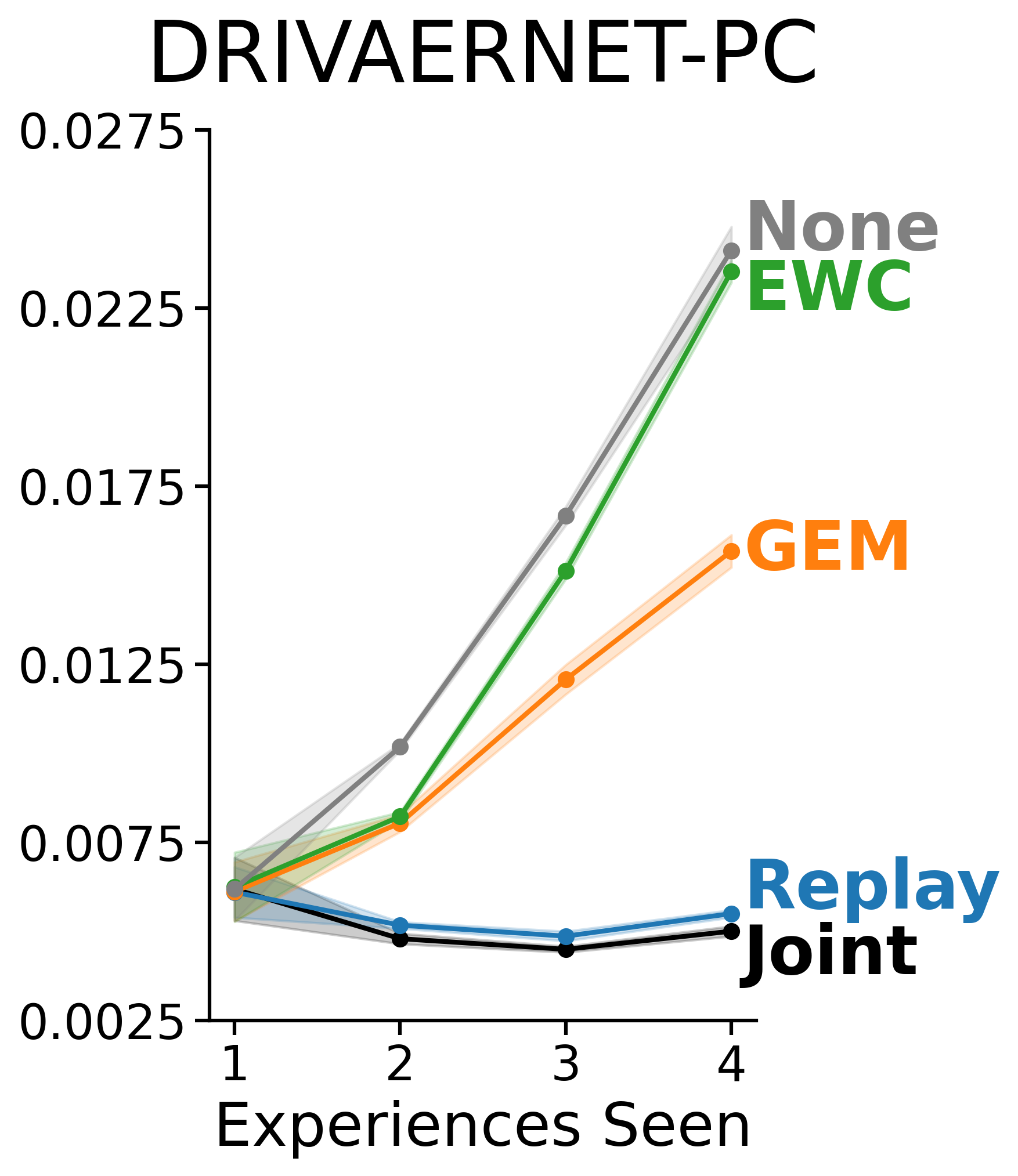}
    \caption{}
    \label{fig:DRIVAERNET-PC_mae}
  \end{subfigure}

  \vspace{0.4cm}

  \begin{subfigure}[b]{0.19\textwidth}
    \includegraphics[width=\textwidth]{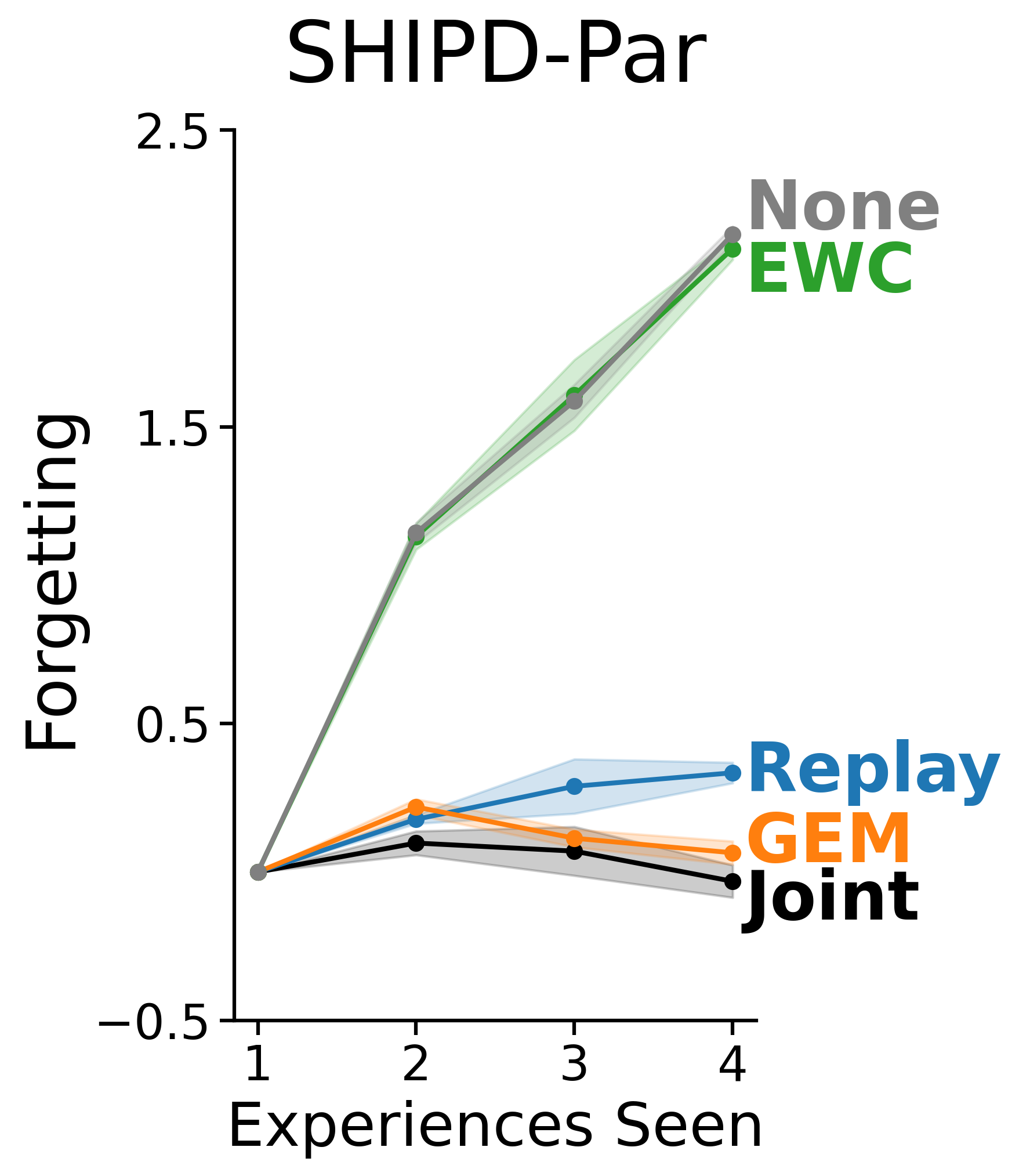}
    \caption{}
    \label{fig:SHIPDsub-Par_forgetting}
  \end{subfigure}
  \begin{subfigure}[b]{0.19\textwidth}
    \includegraphics[width=\textwidth]{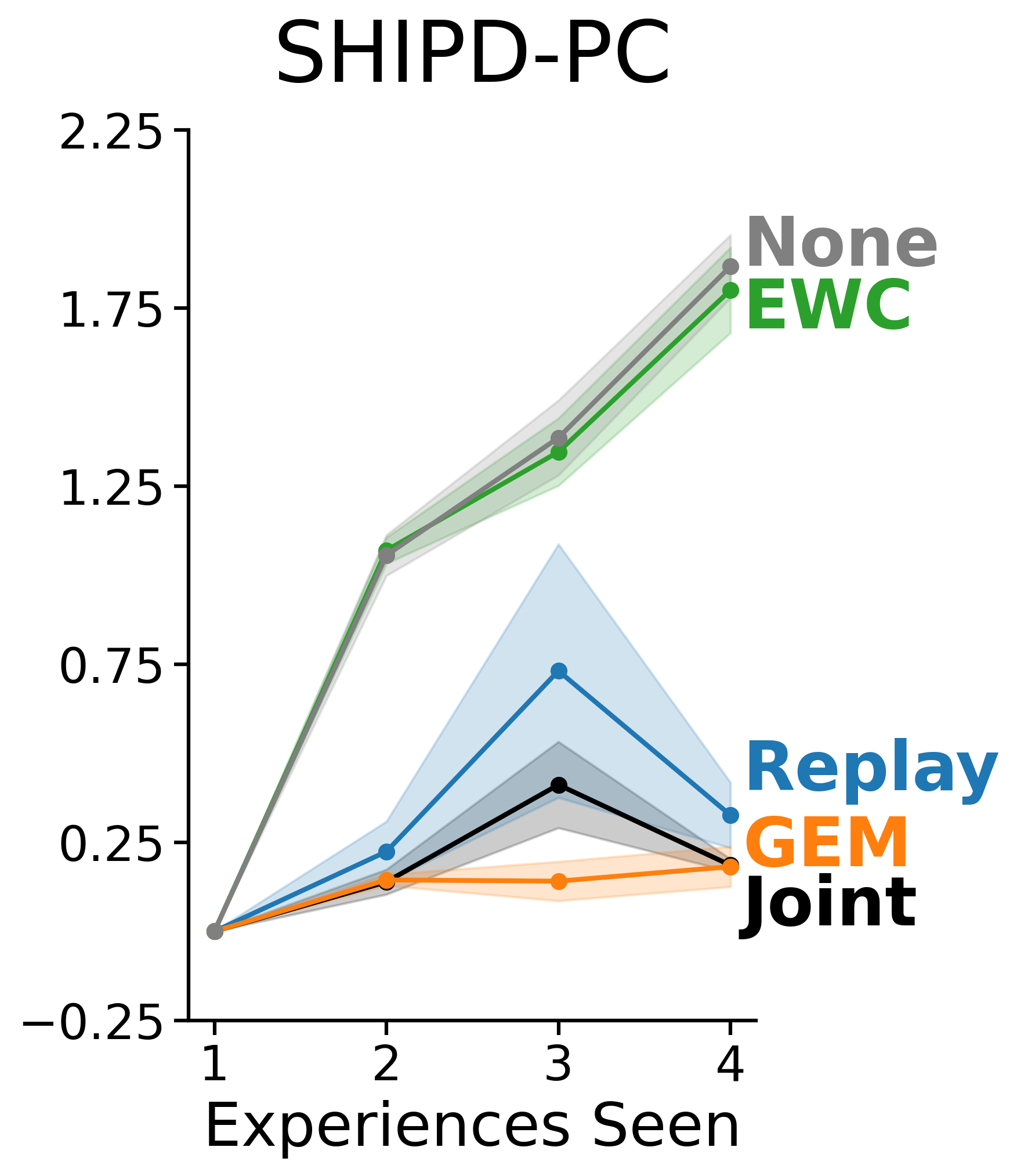}
    \caption{}
    \label{fig:SHIPD-PC_forgetting}
  \end{subfigure}
  \begin{subfigure}[b]{0.19\textwidth}
    \includegraphics[width=\textwidth]{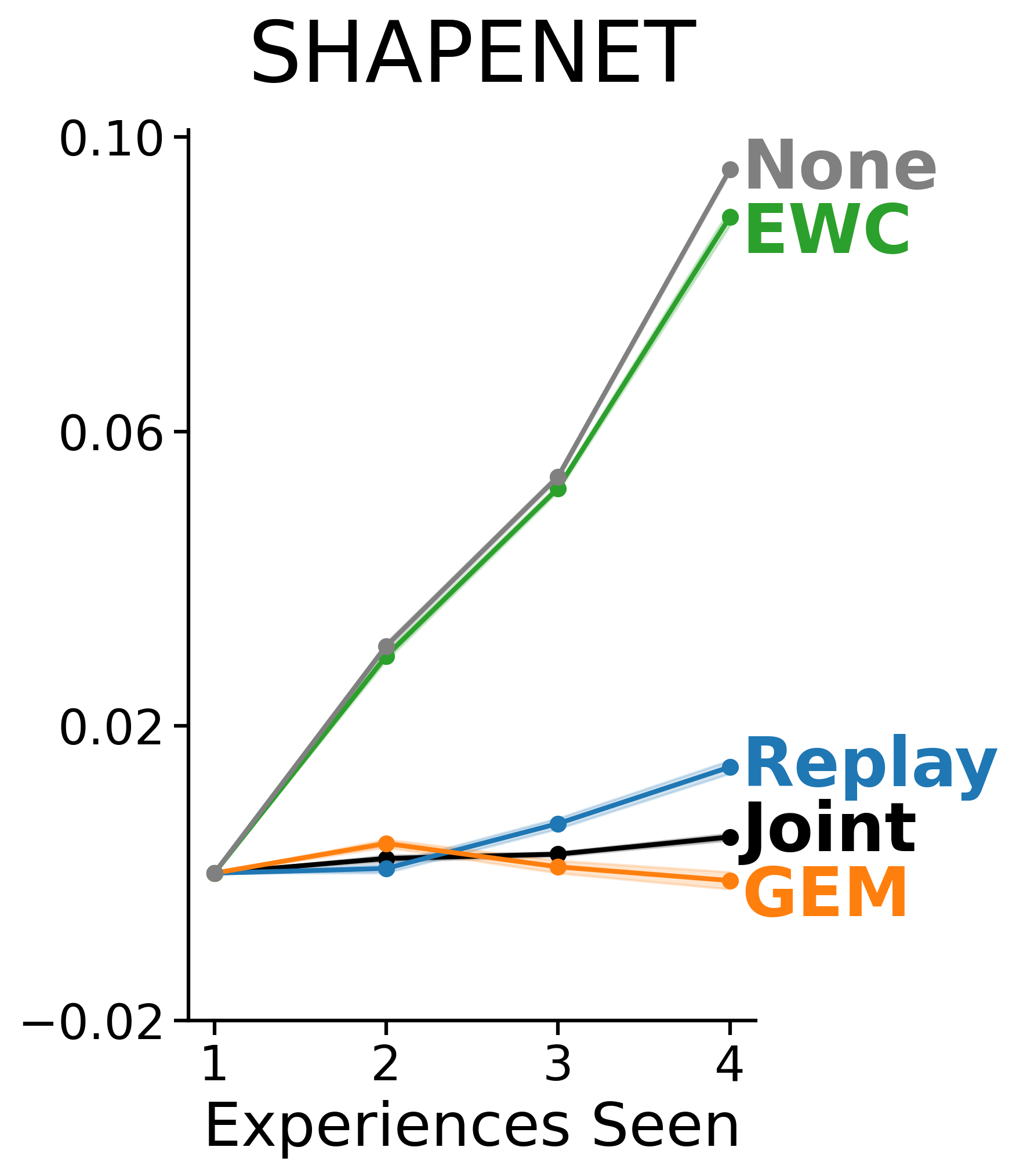}
    \caption{}
    \label{fig:SHAPENET-PC_forgetting}
  \end{subfigure}
  \begin{subfigure}[b]{0.19\textwidth}
    \includegraphics[width=\textwidth]{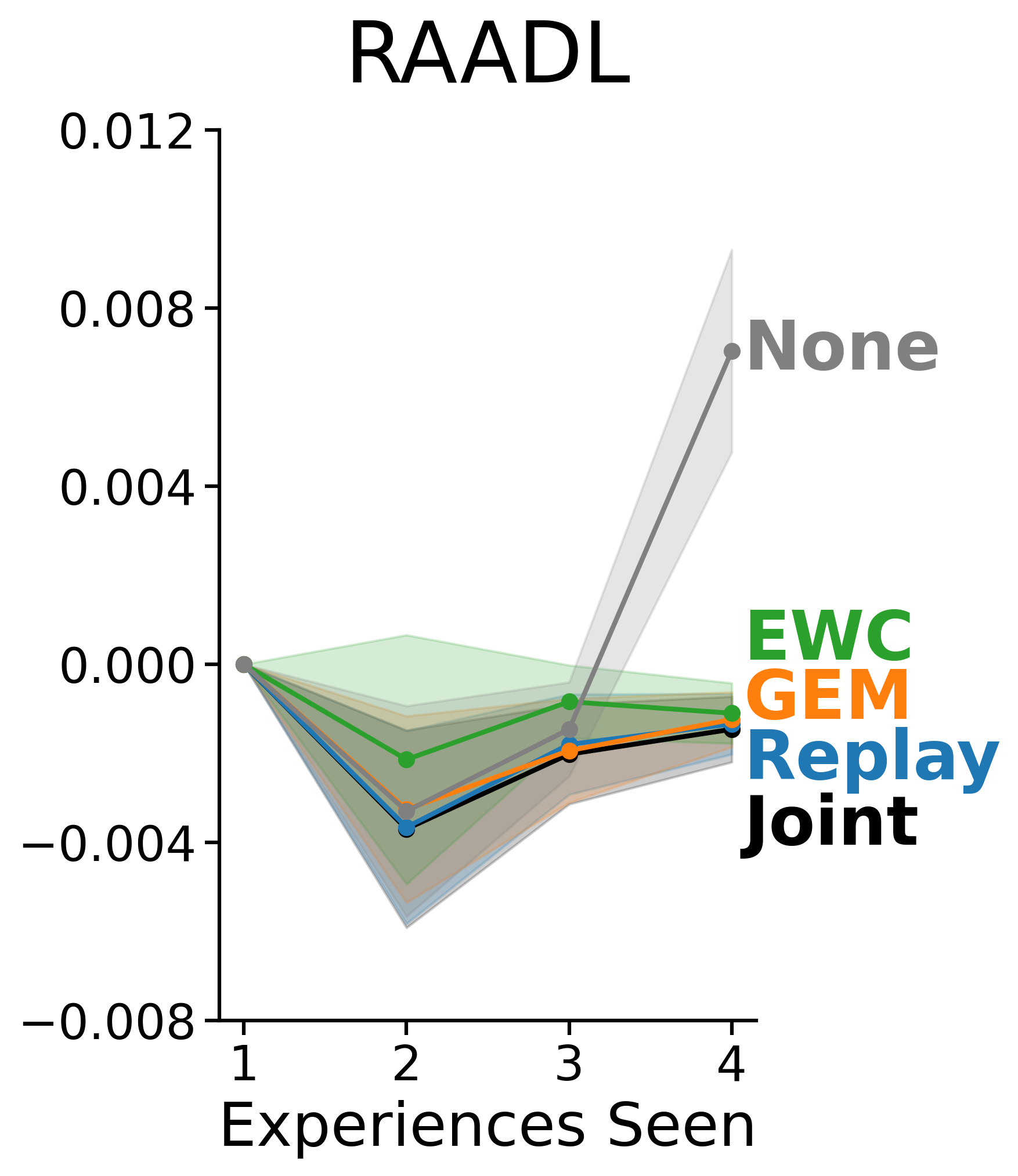}
    \caption{}
    \label{fig:RAADL-PC_forgetting}
  \end{subfigure}
  \begin{subfigure}[b]{0.19\textwidth}
    \includegraphics[width=\textwidth]{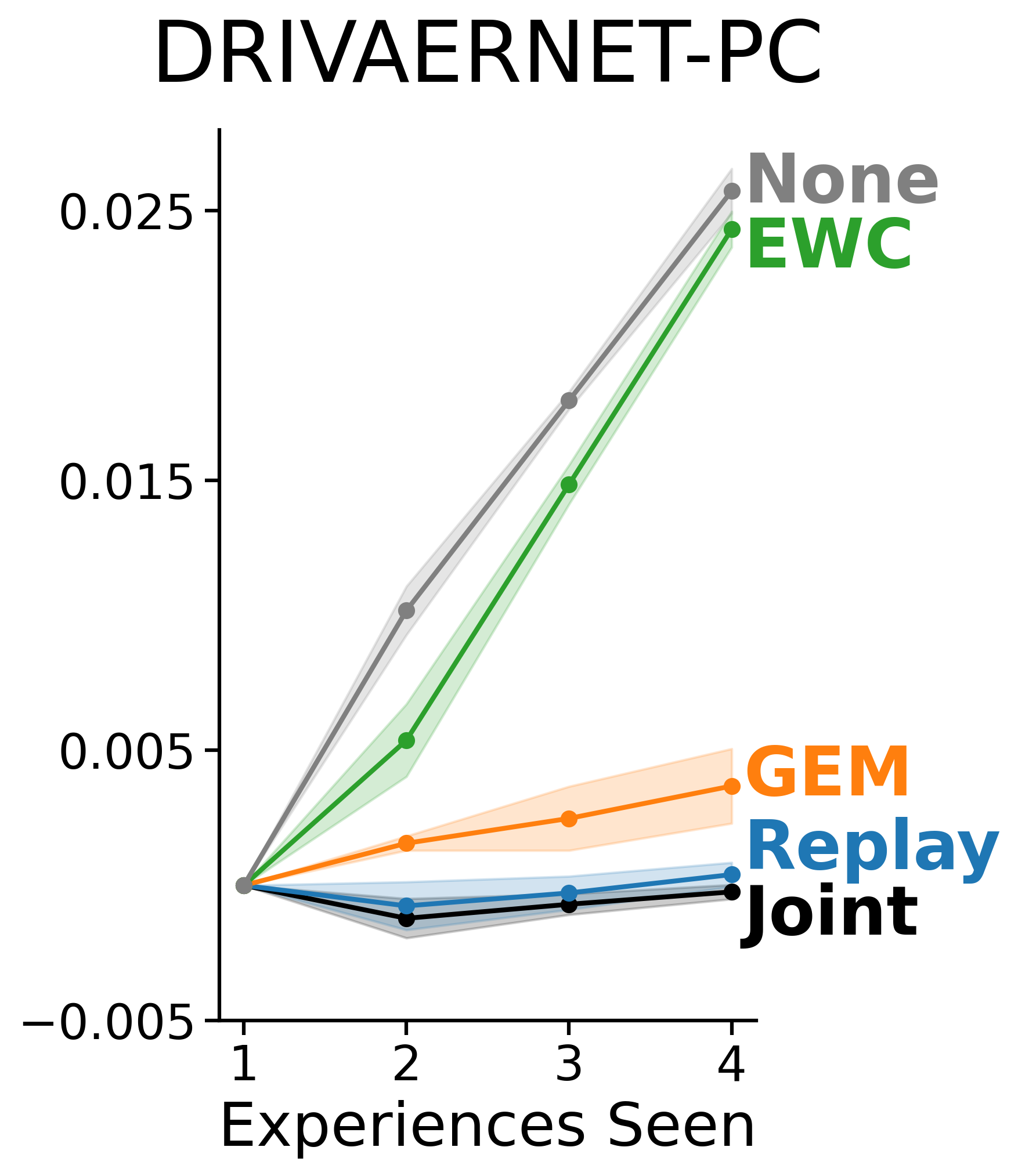}
    \caption{}
    \label{fig:DRIVAERNET-PC_forgetting}
  \end{subfigure}

  \vspace{0.6cm}

  \begin{subfigure}[b]{0.2\textwidth}
    \includegraphics[width=\textwidth]{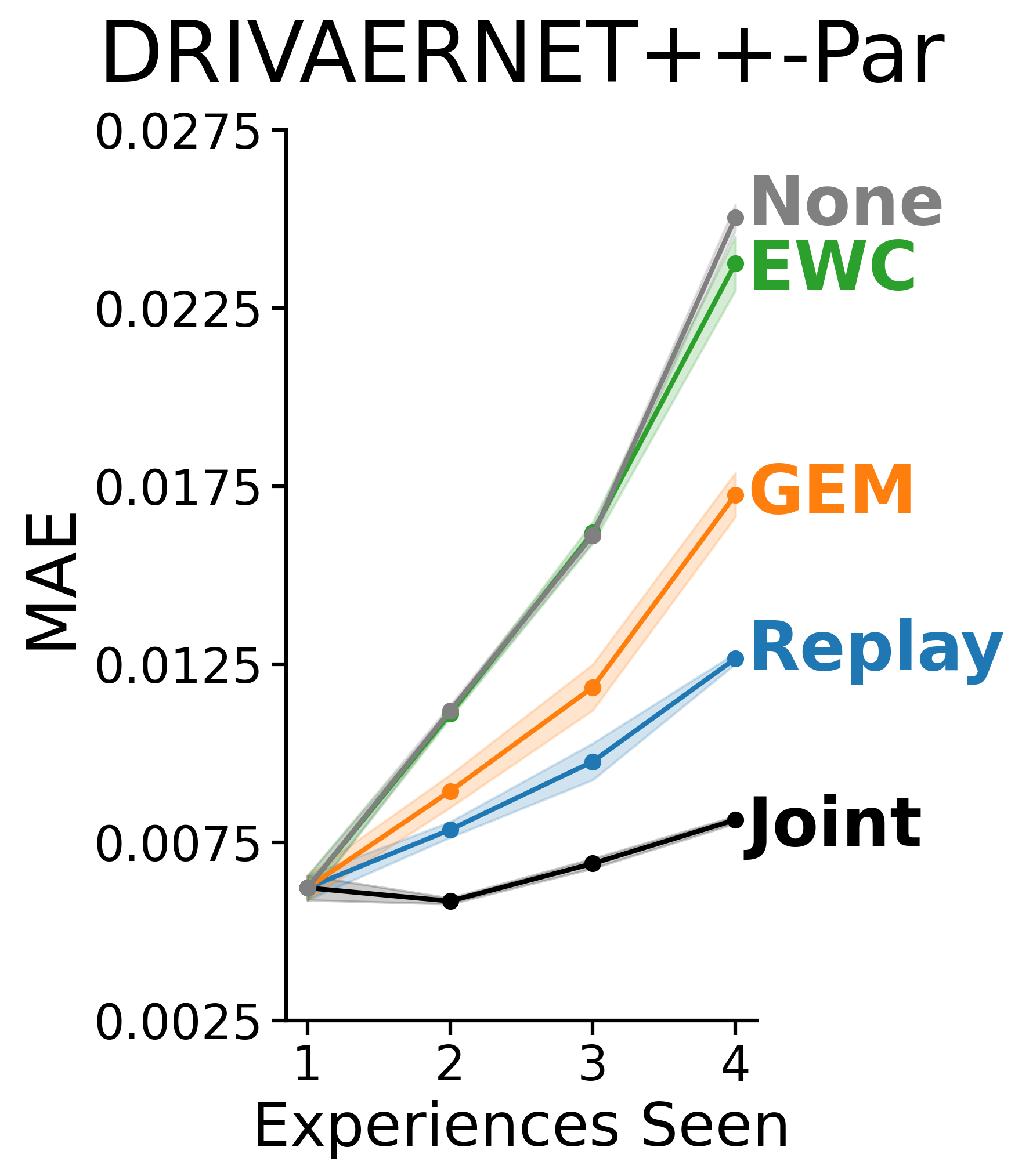}
    \caption{}
    \label{fig:DRIVAERNETplusplus-Par_mae}
  \end{subfigure}
  \begin{subfigure}[b]{0.2\textwidth}
    \includegraphics[width=\textwidth]{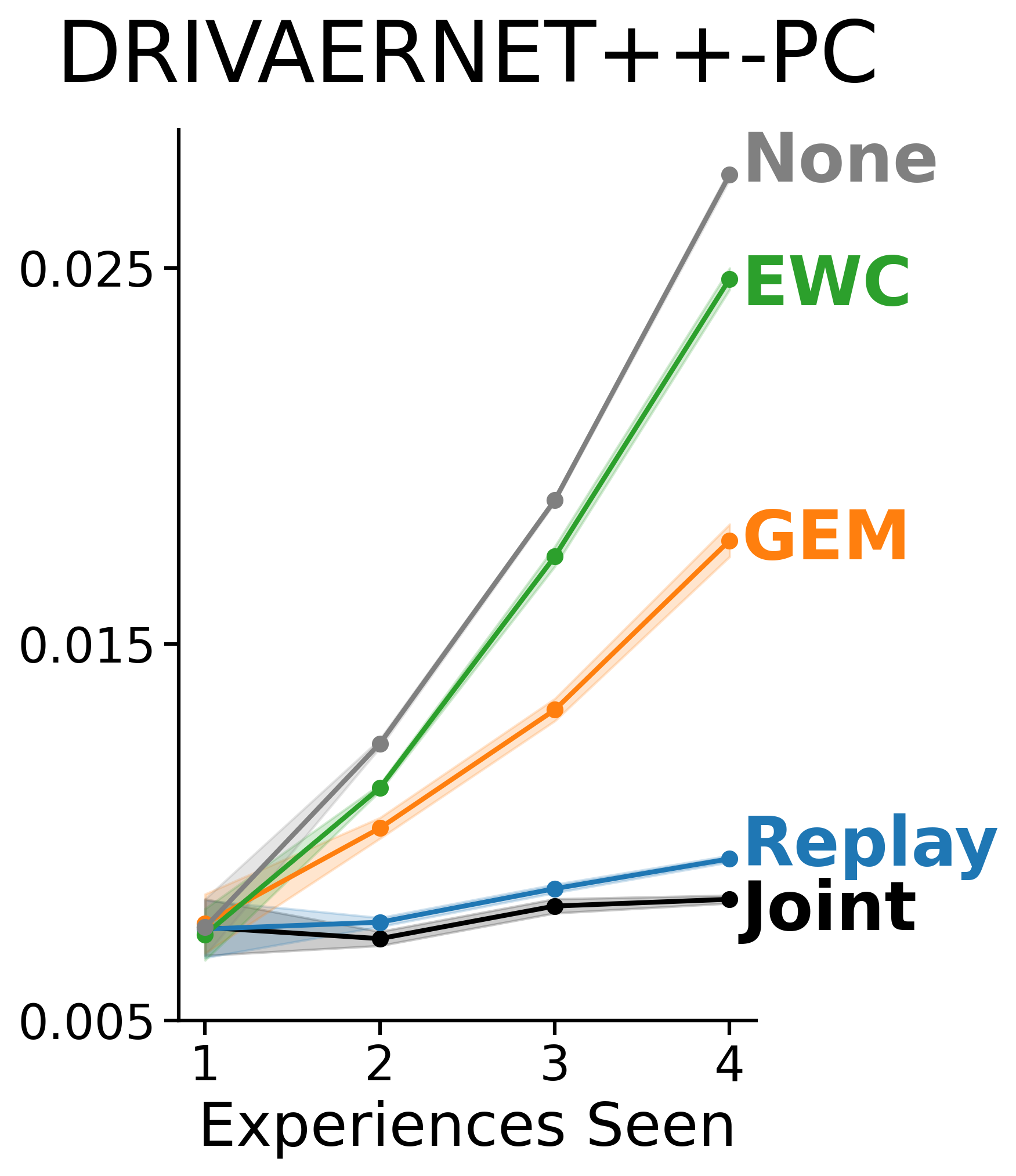}
    \caption{}
    \label{fig:DRIVAERNETplusplus-PC_mae}
  \end{subfigure}
  \begin{subfigure}[b]{0.2\textwidth}
    \includegraphics[width=\textwidth]{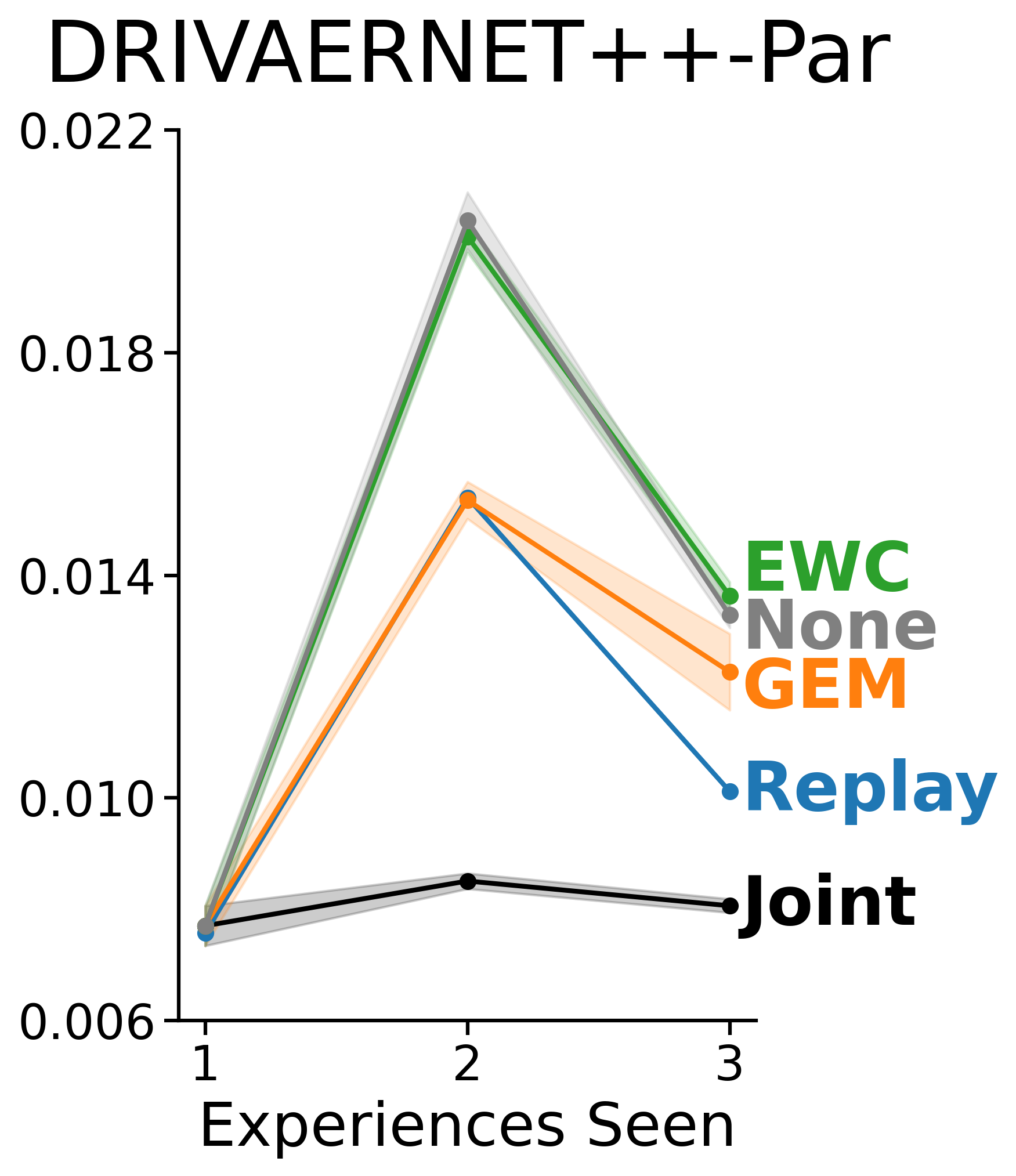}
    \caption{Input Inc.}
    \label{fig:DRIVAERNETplusplus-Par_input_mae}
  \end{subfigure}
  \begin{subfigure}[b]{0.2\textwidth}
    \includegraphics[width=\textwidth]{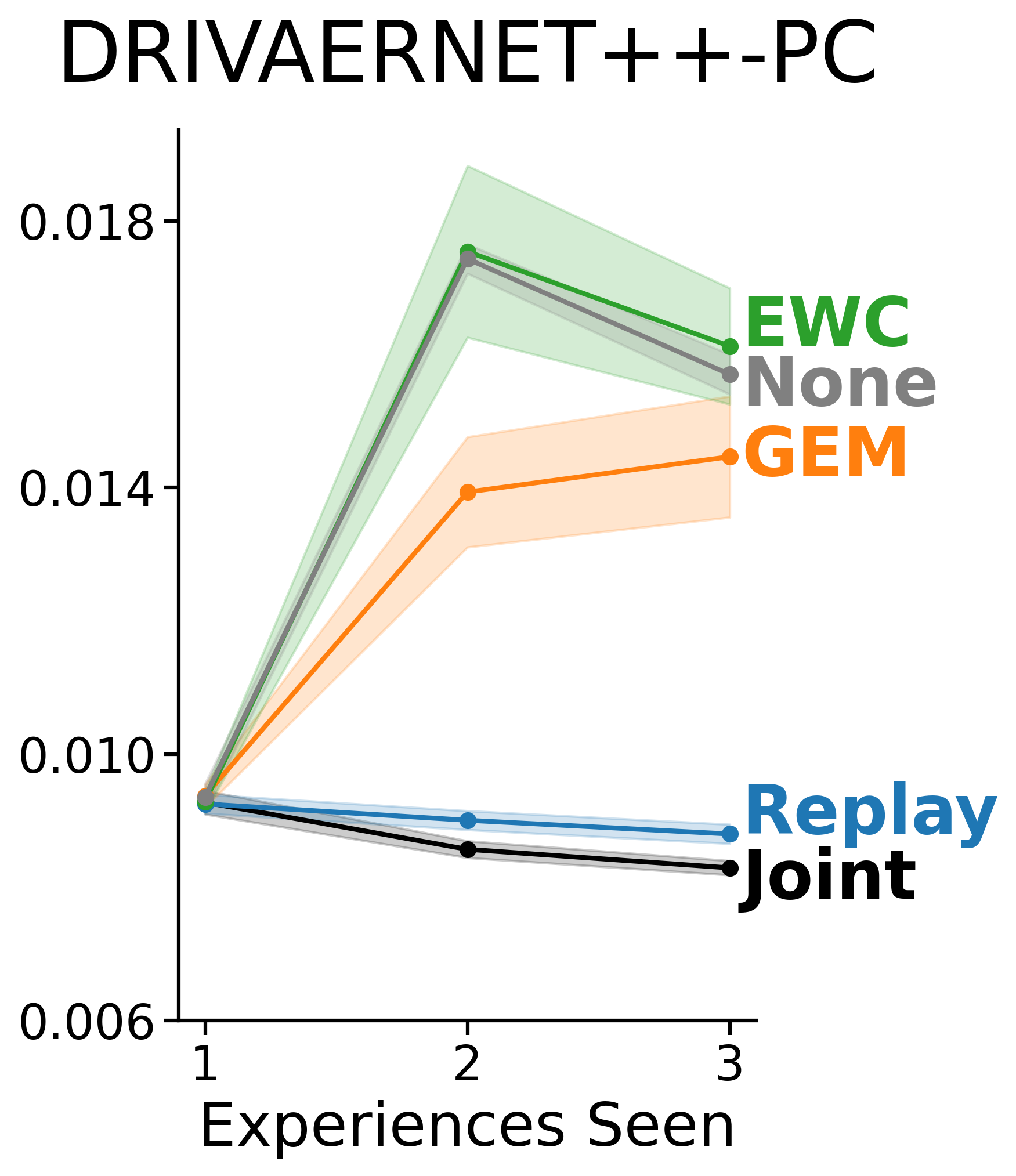}
    \caption{Input Inc.}
    \label{fig:DRIVAERNETplusplus-PC_input_mae}
  \end{subfigure}

  \vspace{0.4cm}

  \begin{subfigure}[b]{0.19\textwidth}
    \includegraphics[width=\textwidth]{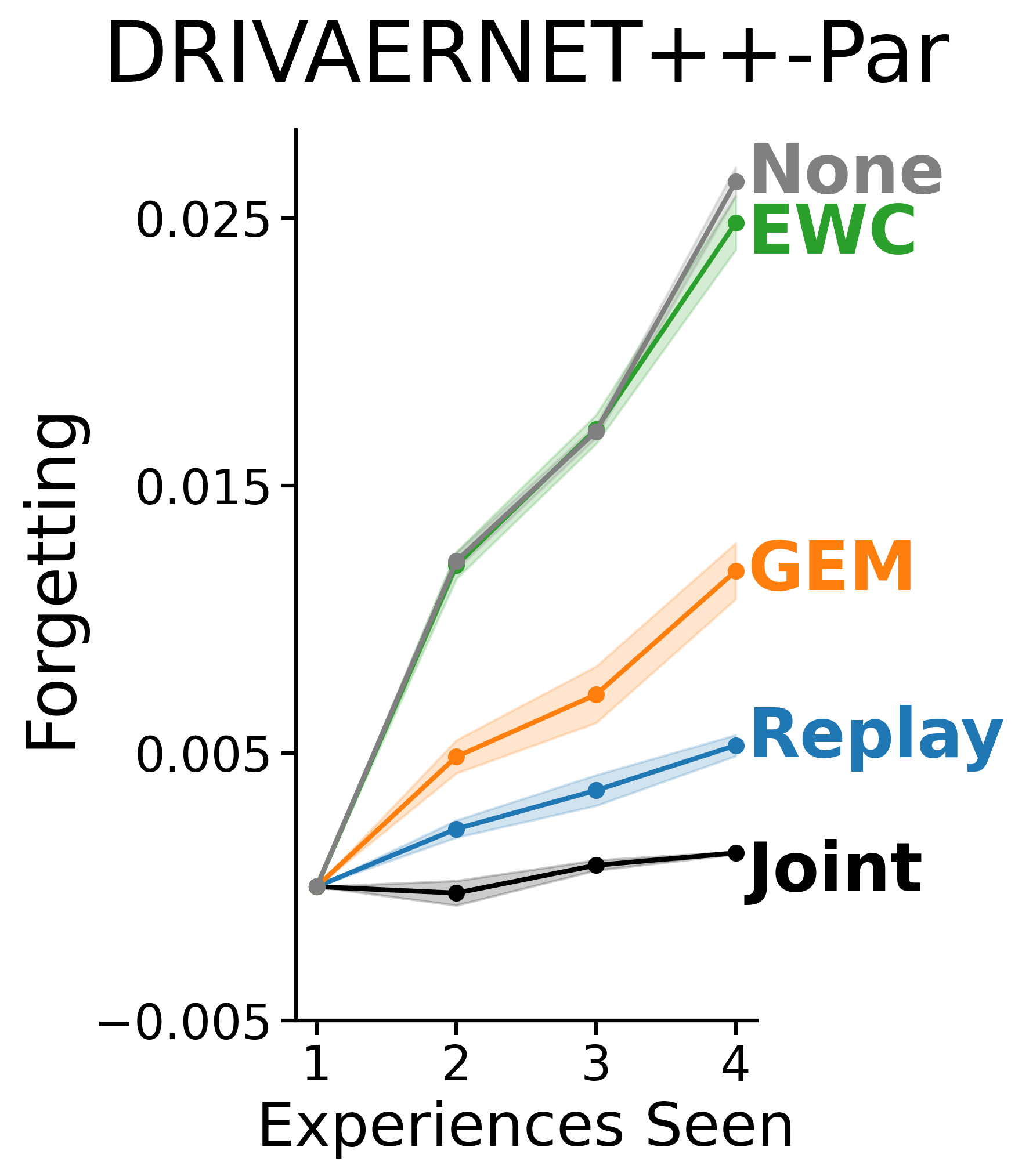}
    \caption{}
    \label{fig:DRIVAERNETplusplus-Par_forgetting}
  \end{subfigure}
  \begin{subfigure}[b]{0.19\textwidth}
    \includegraphics[width=\textwidth]{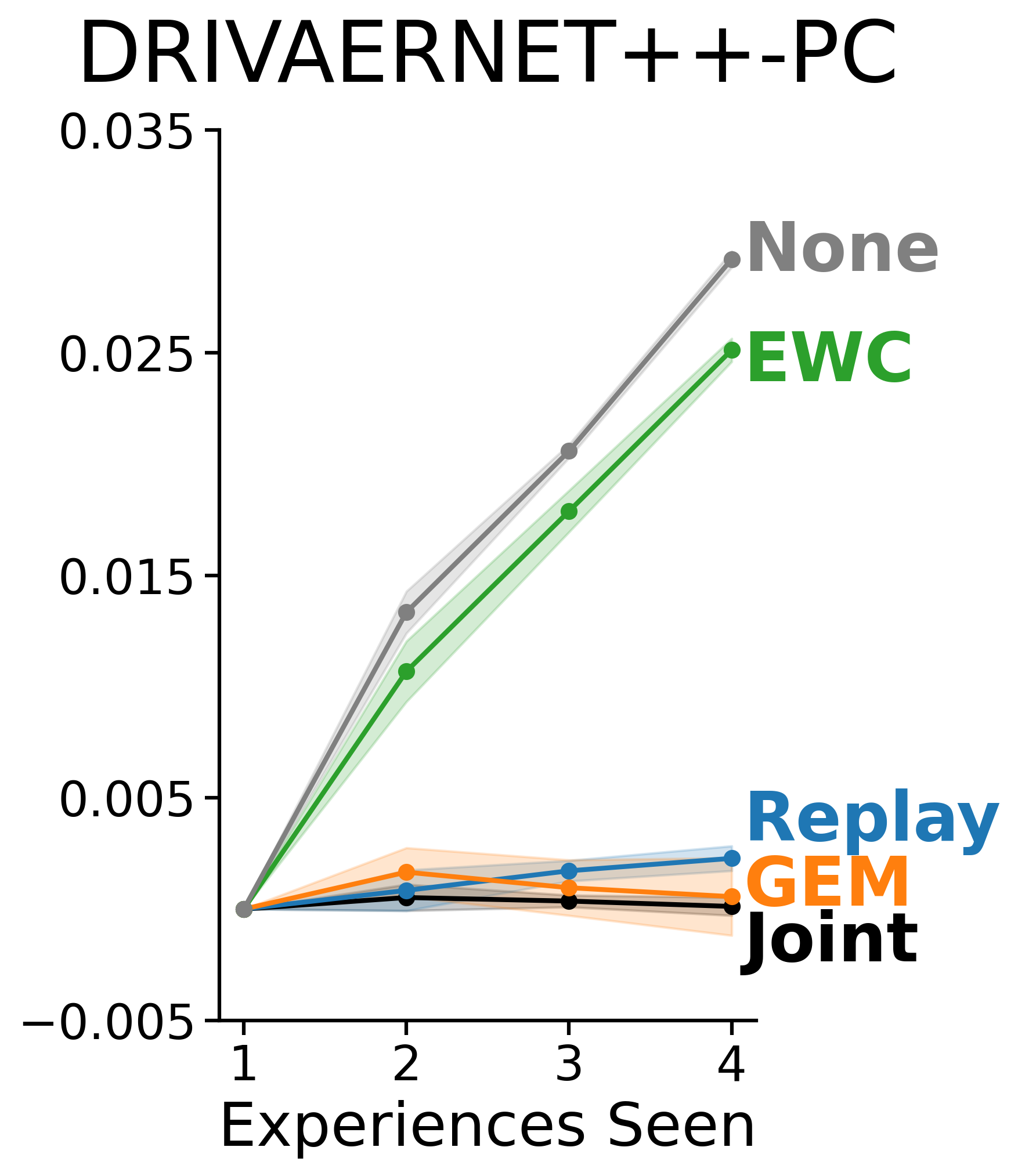}
    \caption{}
    \label{fig:DRIVAERNETplusplus-PC_forgetting}
  \end{subfigure}
  \begin{subfigure}[b]{0.19\textwidth}
    \includegraphics[width=\textwidth]{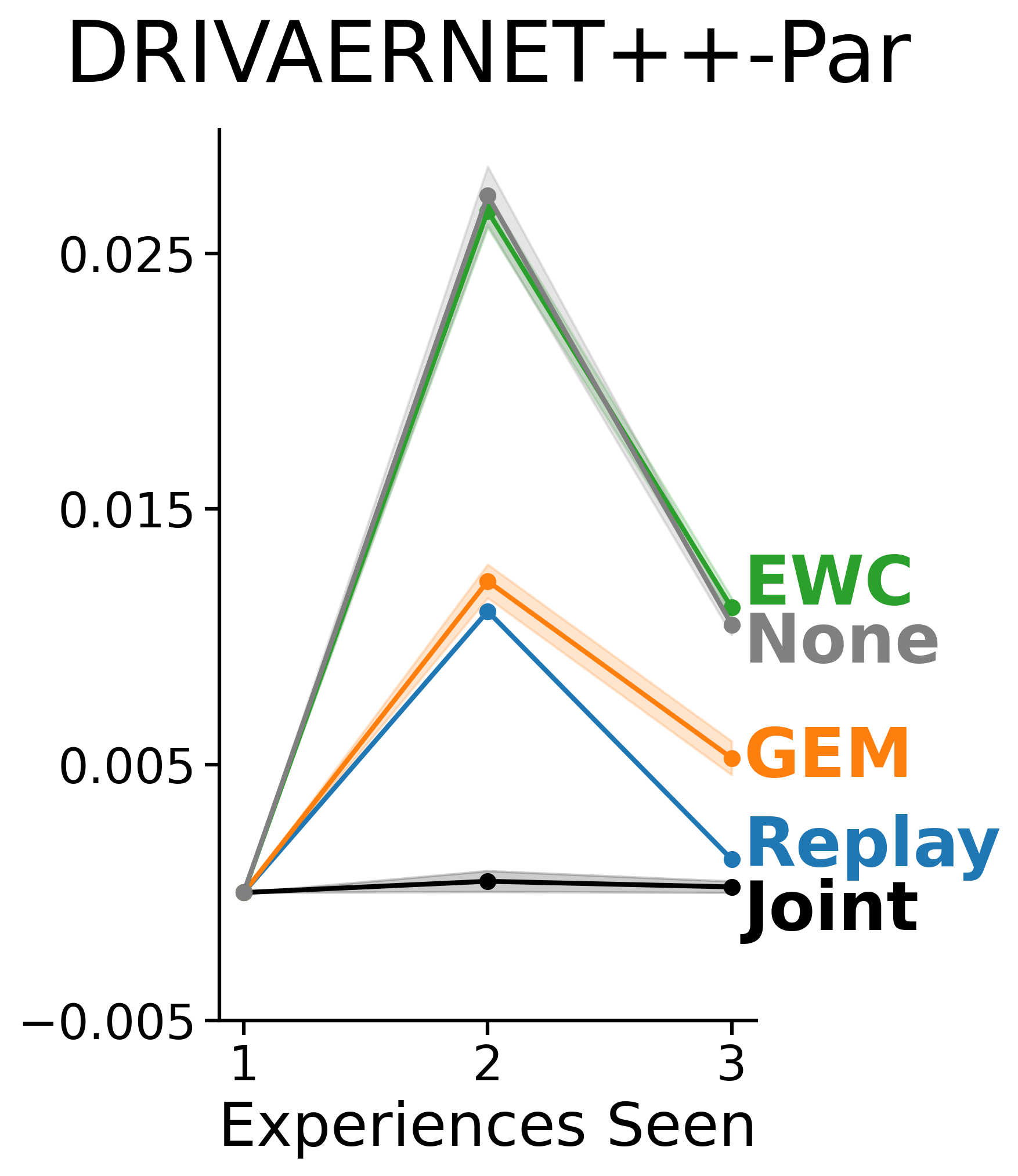}
    \caption{Input Inc.}
    \label{fig:DRIVAERNETplusplus-Par_input_forgetting}
  \end{subfigure}
  \begin{subfigure}[b]{0.19\textwidth}
    \includegraphics[width=\textwidth]{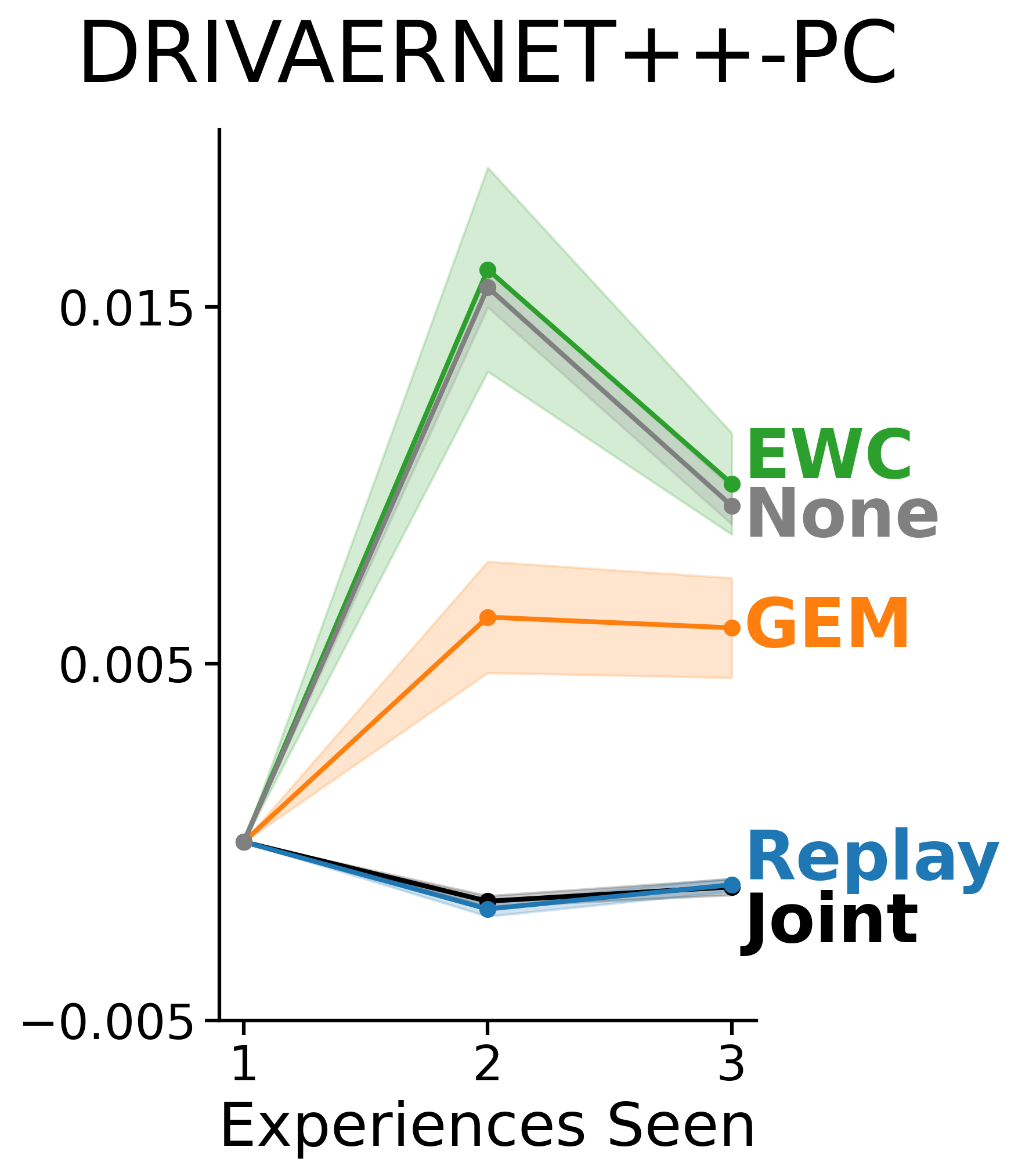}
    \caption{Input Inc.}
    \label{fig:DRIVAERNETplusplus-PC_input_forgetting}
  \end{subfigure}

  \caption{Incremental MAE and Forgetting across all benchmarks, including both bin-incremental and input-incremental (indicated under plot) scenarios.}
  \label{fig:all-benchmarks-grid}
\end{figure*}

\paragraph*{SplitSHIPD-Par}
Replay achieved the lowest final MPE of 11.51\%, outperforming all continual strategies but trailing behind the Joint strategy’s 8.11\% on the full test set. As shown in Figure \ref{fig:SHIPDsub-Par_mae}, Replay also maintained the lowest incremental MAE across experiences. Despite GEM achieving the best final MAE on experiences 2 and 3, and EWC on experience 4 (see Table \ref{tab:comparison_splitshipPar}), these gains did not translate into a lower overall error. Replay’s consistently strong performance across all experiences enabled it to retain the overall lead in final performance. Figure \ref{fig:SHIPDsub-Par_forgetting} further illustrates that GEM achieved lower forgetting incrementally than Replay, yet this did not result in a superior prediction accuracy. Due to the low-dimensional nature of the parametric representation, all strategies showed a reduction in runtime compared to the Joint baseline, with EWC achieving a 36\% reduction, followed by GEM's 26\% reduction and Replay's 23\% reduction.

\begin{table*}[h]
    \centering
        \small
    \renewcommand{\arraystretch}{0.9}
    \setlength{\aboverulesep}{0.5ex}
    \setlength{\belowrulesep}{0.5ex}
    \resizebox{\textwidth}{!}{%
    \begin{tabular}{l *{4}{c} @{\hspace{2em}} *{3}{c}}
        \toprule
        \multicolumn{1}{c}{\textbf{Metrics}} & \multicolumn{4}{c}{\textbf{MAE ($\times 10^{-3}$)} ↓} & \multicolumn{3}{c}{\textbf{Forgetting Ratio} ↓} \\
        \midrule
        \multicolumn{1}{c}{\textbf{Experience}} & \textbf{1} & \textbf{2} & \textbf{3} & \textbf{4} & \textbf{1} & \textbf{2} & \textbf{3} \\
        \midrule
        Baseline & $0.59 \pm 0.04$ & $0.49 \pm 0.05$ & $0.49 \pm 0.03$ & $0.71 \pm 0.04$  & $0.06 \pm 0.07$ & $-0.03 \pm 0.21$ & $-0.15 \pm 0.19$ \\
        Naive & 
        $4.07 \pm 0.04$ & $2.29 \pm 0.03$ & $1.29 \pm 0.03$ & $0.53 \pm 0.02$  & 
        $6.26 \pm 0.38$ & $6.09 \pm 0.28$ & $3.18 \pm 0.37$ \\
        \midrule
        Replay & 
        $\mathbf{1.03 \pm 0.08}$ & $0.82 \pm 0.04$ & $0.74 \pm 0.02$ & $0.68 \pm 0.06$ & 
        $\mathbf{0.84 \pm 0.12}$ & $0.72 \pm 0.10$ & $0.39 \pm 0.32$ \\
        GEM & 
        $1.18 \pm 0.04$ & $\mathbf{0.71 \pm 0.05}$ & $\mathbf{0.73 \pm 0.05}$ & $1.41 \pm 0.07$ & 
        $1.11 \pm 0.08$ & $\mathbf{0.09 \pm 0.06}$ & $\mathbf{-0.39 \pm 0.11}$ \\
        EWC & 
        $4.02 \pm 0.05$ & $2.26 \pm 0.05$ & $1.24 \pm 0.05$ & $\mathbf{0.52 \pm 0.02}$ & 
        $6.20 \pm 0.13$ & $6.04 \pm 0.30$ & $2.65 \pm 0.27$ \\
        \bottomrule
    \end{tabular}%
    }
    \caption{Comparison of final MAE and Forgetting on each experience across different strategies for the SplitSHIPD-Par benchmark in the bin incremental scenario}
    \label{tab:comparison_splitshipPar}
\end{table*}

\paragraph*{SplitSHIPD-PC}
Replay achieved the lowest final MPE of 12.75\%, compared to the Joint strategy's 11.21\%, on the full test set, the lowest overall forgetting ratio of 0.40, and the lowest incremental MAE as the model encountered new experiences (see Figure \ref{fig:SHIPD-PC_mae}).  Table \ref{tab:comparison_splitshipPC} reports each strategy's final MAE and forgetting ratio on each individual experience. In terms of forgetting, GEM achieved the a comparable overall forgetting ratio to Replay of 0.47 and demonstrated lower incremental forgetting than Replay across new experiences (see Figure \ref{fig:SHIPD-PC_forgetting}). Despite this and strong final MAE performance on experiences 2 and 3 (see Table \ref{tab:comparison_splitshipPC}), the corresponding trends seen in Figure \ref{fig:SHIPD-PC_mae} demonstrate that GEM's lower forgetting did not correlate with improved overall or incremental MAE over the Joint or Replay strategies. This phenomenon indicates GEM's lack of adaptability to new experiences. While Table \ref{tab:comparison_splitshipPC} shows the EWC strategy leading to the best final MAE on experience 4, severe forgetting on all previous experiences demonstrates its ineffectiveness on this benchmark. Considering computational efficiency, Replay reduced runtime by 22\% from the Joint baseline, while GEM achieved only a 10\% reduction. EWC substantially reduced train time by 58\%, but was ultimately ineffective as a strategy.

\begin{table*}[h]
    \centering
        \small
    \renewcommand{\arraystretch}{0.9}
    \setlength{\aboverulesep}{0.5ex}
    \setlength{\belowrulesep}{0.5ex}
    \resizebox{\textwidth}{!}{%
    \begin{tabular}{l *{4}{c} @{\hspace{2em}} *{3}{c}}
        \toprule
        \multicolumn{1}{c}{\textbf{Metrics}} & \multicolumn{4}{c}{\textbf{MAE ($\times 10^{-3}$)} ↓} & \multicolumn{3}{c}{\textbf{Forgetting Ratio} ↓} \\
        \midrule
        \multicolumn{1}{c}{\textbf{Experience}} & \textbf{1} & \textbf{2} & \textbf{3} & \textbf{4} & \textbf{1} & \textbf{2} & \textbf{3} \\
        \midrule
        Cumulative & 
        $0.89 \pm 0.08$ & $0.72 \pm 0.02$ & $0.69 \pm 0.07$ & $0.84 \pm 0.18$ & 
        $0.43 \pm 0.08$ & $0.40 \pm 0.16$ & $0.16 \pm 0.20$ \\
        Naive & 
        $3.79 \pm 0.15$ & $2.04 \pm 0.12$ & $1.09 \pm 0.11$ & $0.55 \pm 0.01$ & 
        $5.03 \pm 0.26$ & $5.12 \pm 0.46$ & $2.06 \pm 0.40$ \\
        \midrule
        Replay & 
        $\mathbf{1.11 \pm 0.11}$ & $0.88 \pm 0.07$ & $0.71 \pm 0.02$ & $0.81 \pm 0.07$ & 
        $\mathbf{0.84 \pm 0.23}$ & $0.76 \pm 0.46$ & $0.33 \pm 0.50$ \\
        GEM & 
        $2.21 \pm 0.08$ & $\mathbf{0.61 \pm 0.08}$ & $\mathbf{0.48 \pm 0.07}$ & $1.58 \pm 0.15$ & 
        $2.58 \pm 0.13$ & $\mathbf{-0.40 \pm 0.07}$ & $\mathbf{-0.57 \pm 0.08}$ \\
        EWC & 
        $3.74 \pm 0.19$ & $1.97 \pm 0.16$ & $0.94 \pm 0.15$ & $\mathbf{0.55 \pm 0.08}$ & 
        $5.05 \pm 0.35$ & $5.45 \pm 0.46$ & $1.87 \pm 0.56$ \\
        \bottomrule
    \end{tabular}%
    }
    \caption{Comparison of final MAE and Forgetting on each experience across different strategies for the SplitSHIPD-PC benchmark in the bin incremental scenario}
    \label{tab:comparison_splitshipPC}
\end{table*}

\paragraph*{SplitSHAPENET}
Replay achieved the lowest final MPE, outperforming all other strategies while remaining within 1.5 percentage points of the Joint baseline, as seen in Table \ref{tab:benchmark_comparison}. Replay also maintained the lowest incremental MAE as the model encountered new experiences (see Figure \ref{fig:SHAPENET-PC_mae}). Table \ref{tab:comparison_splitshapenet} reports each strategy's final MAE and forgetting ratio across individual experiences, in which GEM and EWC achieved slightly higher MAEs on experiences 2 and 4, respectively. GEM also achieved the best overall forgetting ratio and exhibited lower incremental forgetting than Replay, even dipping below the Joint strategy (see Figure \ref{fig:SHAPENET-PC_forgetting}). However, these slight improvements on earlier experiences did not translate to better performance than Replay on overall MAE or MPE, highlighting GEM’s limited adaptability to new experiences. EWC performed comparably to the Naive baseline in terms of accuracy, further demonstrating its inability to mitigate catastrophic forgetting on this benchmark. In terms of runtime, GEM required the most training time among continual strategies, only reducing train time by 7\% from the Joint baseline, while Replay achieved strong performance with an approximately 45\% train time reduction.

\begin{table*}[h]
    \centering
    \small
    \renewcommand{\arraystretch}{0.9}
    \setlength{\aboverulesep}{0.5ex}
    \setlength{\belowrulesep}{0.5ex}
    \resizebox{\textwidth}{!}{%
    \begin{tabular}{l *{4}{c} @{\hspace{2em}} *{3}{c}}
        \toprule
        \multicolumn{1}{c}{\textbf{Metrics}} & \multicolumn{4}{c}{\textbf{MAE ($\times 10^{-3}$)} ↓} & \multicolumn{3}{c}{\textbf{Forgetting Ratio} ↓} \\
        \midrule
        \multicolumn{1}{c}{\textbf{Experience}} & \textbf{1} & \textbf{2} & \textbf{3} & \textbf{4} & \textbf{1} & \textbf{2} & \textbf{3} \\
        \midrule
            Cumulative & 
            $0.0246 \pm 0.0014$ & $0.0187 \pm 0.0008$ & $0.0201 \pm 0.0007$ & $0.0386 \pm 0.0018$ & 
            $0.86 \pm 0.11$ & $0.34 \pm 0.06$ & $-0.06 \pm 0.04$ \\
            Naive & 
            $0.1488 \pm 0.0011$ & $0.1045 \pm 0.0008$ & $0.0646 \pm 0.0009$ & $0.0249 \pm 0.0013$ & 
            $10.28 \pm 0.85$ & $12.18 \pm 0.26$ & $5.26 \pm 0.17$ \\
            \midrule
            Replay & 
            $\mathbf{0.0326 \pm 0.0024}$ & $0.0300 \pm 0.0015$ & $\mathbf{0.0260 \pm 0.0011}$ & $0.0288 \pm 0.0014$ & 
            $\mathbf{1.36 \pm 0.21}$ & $1.07 \pm 0.18$ & $0.53 \pm 0.06$ \\
            GEM & 
            $0.0454 \pm 0.0045$ & $\mathbf{0.0117 \pm 0.0010}$ & $0.0419 \pm 0.0025$ & $0.1232 \pm 0.0029$ & 
            $2.20 \pm 0.32$ & $\mathbf{-0.62 \pm 0.05}$ & $\mathbf{-0.27 \pm 0.03}$ \\
            EWC & 
            $0.1436 \pm 0.0013$ & $0.0986 \pm 0.0017$ & $0.0578 \pm 0.0022$ & $\mathbf{0.0227 \pm 0.0005}$ & 
            $9.45 \pm 0.72$ & $11.02 \pm 0.39$ & $4.50 \pm 0.41$ \\
            \bottomrule
        \end{tabular}%
    }
    \caption{Comparison of final MAE and Forgetting on each experience across different strategies for the SplitSHAPENET benchmark in the bin incremental scenario}
    \label{tab:comparison_splitshapenet}
\end{table*}

\paragraph*{SplitRAADL}

\begin{table*}[h]
    \centering
        \small
    \renewcommand{\arraystretch}{0.9}
    \setlength{\aboverulesep}{0.5ex}
    \setlength{\belowrulesep}{0.5ex}
    \resizebox{\textwidth}{!}{%
    \begin{tabular}{l *{4}{c} @{\hspace{2em}} *{3}{c}}
        \toprule
        \multicolumn{1}{c}{\textbf{Metrics}} & \multicolumn{4}{c}{\textbf{MAE ($\times 10^{-3}$)} ↓} & \multicolumn{3}{c}{\textbf{Forgetting Ratio} ↓} \\
        \midrule
        \multicolumn{1}{c}{\textbf{Experience}} & \textbf{1} & \textbf{2} & \textbf{3} & \textbf{4} & \textbf{1} & \textbf{2} & \textbf{3} \\
        \midrule
        Cumulative & 
        $0.330 \pm 0.046$ & $0.343 \pm 0.032$ & $0.390 \pm 0.058$ & $1.423 \pm 0.195$ & 
        $-0.90 \pm 0.09$ & $-0.41 \pm 0.14$ & $-0.20 \pm 0.33$ \\
        Naive & 
        $15.398 \pm 4.968$ & $8.319 \pm 3.209$ & $3.432 \pm 1.596$ & $1.598 \pm 0.270$ & 
        $3.71 \pm 3.99$ & $8.85 \pm 4.89$ & $3.64 \pm 2.48$ \\
        \midrule
        Replay & 
        $\mathbf{0.526 \pm 0.192}$ & $\mathbf{0.572 \pm 0.068}$ & $\mathbf{0.610 \pm 0.047}$ & $\mathbf{1.373 \pm 0.277}$ & 
        $\mathbf{-0.85 \pm 0.11}$ & $-0.12 \pm 0.46$ & $\mathbf{-0.13 \pm 0.16}$ \\
        GEM & 
        $0.843 \pm 0.430$ & $0.871 \pm 0.316$ & $0.676 \pm 0.161$ & $1.575 \pm 0.368$ & 
        $-0.75 \pm 0.15$ & $\mathbf{-0.17 \pm 0.34}$ & $0.13 \pm 0.62$ \\
        EWC & 
        $1.574 \pm 0.459$ & $1.646 \pm 0.471$ & $1.625 \pm 0.723$ & $1.951 \pm 0.664$ & 
        $-0.55 \pm 0.29$ & $0.00 \pm 0.68$ & $-0.05 \pm 0.22$ \\
        \bottomrule
    \end{tabular}%
    }
    \caption{Comparison of final MAE and Forgetting on each experience across different strategies for the SplitRAADL benchmark in the bin incremental scenario}
    \label{tab:comparison_splitraadl}
\end{table*}
 
Replay achieved the lowest final MPE of 6.95\%, outperforming GEM and EWC, which reached 7.75\% and 11.96\% respectively, while also achieving the lowest final MAE on each individual experience (see Table \ref{tab:comparison_splitraadl}). Unlike the other benchmarks, SplitRAADL exhibited consistent backward transfer for all strategies, most notably reflected in the drop in incremental MAE after experience 1 (Figure \ref{fig:RAADL-PC_mae}). This trend indicates that models improved on earlier experiences as new data was introduced. Supporting this, Table \ref{tab:comparison_splitraadl} shows negative forgetting ratios across all strategies, signaling performance gains on past experiences. Replay and GEM both demonstrated low forgetting and closely tracked the Joint baseline (Figure \ref{fig:RAADL-PC_forgetting}). While EWC underperformed in overall MAE, it nonetheless improved over the Naive baseline and showed negative forgetting on experiences 1 and 3, demonstrating better performance than in other benchmarks. In terms of computational efficiency, Replay reduced runtime by 17\% from the Joint baseline, GEM reduced it by 26\% and EWC had a substantial 45\% reduction. The overall variability in performance across experiences, visible in both MAE and forgetting trends, suggests that limited data per experience contributed to unstable learning behavior.

\paragraph*{SplitDRIVAERNET-PC}
Replay achieved the lowest final MPE of 2.07\% on the full test set, along with the lowest MAE on each individual experience (see Table \ref{tab:comparison_splitdrivaernetPC}). Figure \ref{fig:DRIVAERNET-PC_mae} shows that Replay consistently maintained the lowest incremental MAE as new experiences were introduced. In contrast, EWC showed substantial forgetting across experiences, with performance comparable to or worse than the Naive baseline. GEM outperformed EWC but still exhibited significant forgetting on the first experience. However, GEM achieved negative forgetting ratios on experiences 2 and 3 (Table \ref{tab:comparison_splitdrivaernetPC}), with a lower reported forgetting raio on experience 3 than Replay. Overall, however, the trend visible in Figure \ref{fig:DRIVAERNET-PC_forgetting} shows Replay experiencing lower incremental forgetting than GEM. Furthermore, while GEM achieved a 13\% reduction in runtime from the Joint strategy, Replay reduced the runtime by approximately 45\%. EWC was the fastest with a 51\% reduction, yet failed to improve over the Naive baseline, highlighting its limited effectiveness on this benchmark.

\begin{table*}[h]
    \centering
        \small
    \renewcommand{\arraystretch}{0.9}
    \setlength{\aboverulesep}{0.5ex}
    \setlength{\belowrulesep}{0.5ex}
    \resizebox{\textwidth}{!}{%
    \begin{tabular}{l *{4}{c} @{\hspace{2em}} *{3}{c}}
        \toprule
        \multicolumn{1}{c}{\textbf{Metrics}} & \multicolumn{4}{c}{\textbf{MAE ($\times 10^{-3}$)} ↓} & \multicolumn{3}{c}{\textbf{Forgetting Ratio} ↓} \\
        \midrule
        \multicolumn{1}{c}{\textbf{Experience}} & \textbf{1} & \textbf{2} & \textbf{3} & \textbf{4} & \textbf{1} & \textbf{2} & \textbf{3} \\
        \midrule
        Cumulative & 
        $0.495 \pm 0.051$ & $0.447 \pm 0.021$ & $0.474 \pm 0.034$ & $0.585 \pm 0.049$ & 
        $-0.19 \pm 0.14$ & $-0.03 \pm 0.08$ & $0.16 \pm 0.10$ \\
        Naive & 
        $4.622 \pm 0.210$ & $3.022 \pm 0.130$ & $1.487 \pm 0.046$ & $0.518 \pm 0.019$ & 
        $6.61 \pm 1.40$ & $6.55 \pm 0.38$ & $2.81 \pm 0.08$ \\
        \midrule
        Replay & 
        $\mathbf{0.601 \pm 0.043}$ & $\mathbf{0.464 \pm 0.038}$ & $\mathbf{0.571 \pm 0.047}$ & $\mathbf{0.563 \pm 0.024}$ & 
        $\mathbf{0.00 \pm 0.22}$ & $\mathbf{-0.07 \pm 0.09}$ & $0.40 \pm 0.09$ \\
        GEM & 
        $2.606 \pm 0.327$ & $0.803 \pm 0.209$ & $0.817 \pm 0.215$ & $2.044 \pm 0.297$ & 
        $3.32 \pm 0.84$ & $-0.03 \pm 0.35$ & $\mathbf{-0.51 \pm 0.11}$ \\
        EWC & 
        $4.704 \pm 0.184$ & $2.913 \pm 0.067$ & $1.240 \pm 0.040$ & $0.557 \pm 0.039$ & 
        $6.71 \pm 2.12$ & $5.09 \pm 0.93$ & $1.70 \pm 0.26$ \\
        \bottomrule
    \end{tabular}%
    }
    \caption{Comparison of final MAE and Forgetting on each experience across different strategies for the SplitDRIVAERNET-PC benchmark in the bin incremental scenario}
    \label{tab:comparison_splitdrivaernetPC}
\end{table*}

\paragraph*{SplitDRIVAERNET++-Par}

\begin{table*}[h]
    \centering
        \small
    \renewcommand{\arraystretch}{0.9}
    \setlength{\aboverulesep}{0.5ex}
    \setlength{\belowrulesep}{0.5ex}
    \resizebox{\textwidth}{!}{%
    \begin{tabular}{l *{4}{c} @{\hspace{2em}} *{3}{c}}
        \toprule
        \multicolumn{1}{c}{\textbf{Metrics}} & \multicolumn{4}{c}{\textbf{MAE ($\times 10^{-3}$)} ↓} & \multicolumn{3}{c}{\textbf{Forgetting Ratio} ↓} \\
        \midrule
        \multicolumn{1}{c}{\textbf{Experience}} & \textbf{1} & \textbf{2} & \textbf{3} & \textbf{4} & \textbf{1} & \textbf{2} & \textbf{3} \\
        \midrule
        Cumulative & 
        $0.0075 \pm 0.0006$ & $0.0076 \pm 0.0003$ & $0.0077 \pm 0.0003$ & $0.0097 \pm 0.0006$ & 
        $0.21 \pm 0.05$ & $0.34 \pm 0.09$ & $0.08 \pm 0.09$ \\
        Naive & 
        $0.0468 \pm 0.0011$ & $0.0304 \pm 0.0005$ & $0.0162 \pm 0.0005$ & $0.0068 \pm 0.0002$ & 
        $6.54 \pm 0.66$ & $6.63 \pm 0.34$ & $2.95 \pm 0.29$ \\
        \midrule
        Replay & 
        $\mathbf{0.0144 \pm 0.0013}$ & $\mathbf{0.0128 \pm 0.0004}$ & $\mathbf{0.0108 \pm 0.0002}$ & $0.0127 \pm 0.0013$ & 
        $\mathbf{1.31 \pm 0.22}$ & $\mathbf{0.74 \pm 0.05}$ & $\mathbf{0.27 \pm 0.11}$ \\
        GEM & 
        $0.0271 \pm 0.0027$ & $0.0174 \pm 0.0014$ & $0.0120 \pm 0.0004$ & $0.0125 \pm 0.0007$ & 
        $3.37 \pm 0.57$ & $1.57 \pm 0.38$ & $0.49 \pm 0.15$ \\
        EWC & 
        $0.0446 \pm 0.0016$ & $0.0289 \pm 0.0014$ & $0.0154 \pm 0.0011$ & $\mathbf{0.0061 \pm 0.0003}$ & 
        $6.20 \pm 0.62$ & $6.25 \pm 0.57$ & $2.68 \pm 0.28$ \\
        \bottomrule
    \end{tabular}%
    }
    \caption{Comparison of final MAE and Forgetting on each experience across different strategies for the SplitDRIVAERNET++-Par benchmark in the bin incremental scenario}
    \label{tab:comparison_bin_splitdrivaernet++Par}
\end{table*}

Replay achieved the best final MPE of 4.90\%, compared to the Joint baseline at 3.13\%, and consistently maintained the lowest MAE across all experiences except for experience 4 (see Table \ref{tab:comparison_bin_splitdrivaernet++Par}). As shown in Figure \ref{fig:DRIVAERNETplusplus-Par_mae}, Replay exhibited strong incremental performance, with MAE increasing gradually but remaining well below that of other continual strategies. Replay also achieved the lowest forgetting ratios across all experiences and maintained low forgetting incrementally, as seen in Figure \ref{fig:DRIVAERNETplusplus-Par_forgetting}, highlighting its stability over time. GEM demonstrated moderate performance, ranking between Replay and EWC in both forgetting and MAE, but did not surpass Replay on any metric. Its performance deviated more substantially after experience 3. EWC showed the highest forgetting ratios overall, with values reaching 6.25, and overfit to the final experience, which resulted in the lowest MAE on experience 4. However, its overall performance remained poor, closely resembling that of the Naive baseline. In terms of computation,  Replay dramatically reduced runtime by 59\% from the Joint baseline, outperforming both GEM (43\% reduction) and EWC (48\% reduction).

\paragraph*{SplitDRIVAERNET++-PC}
\begin{table*}[h]
    \centering
        \small
    \renewcommand{\arraystretch}{0.9}
    \setlength{\aboverulesep}{0.5ex}
    \setlength{\belowrulesep}{0.5ex}
    \resizebox{\textwidth}{!}{%
    \begin{tabular}{l *{4}{c} @{\hspace{2em}} *{3}{c}}
        \toprule
        \multicolumn{1}{c}{\textbf{Metrics}} & \multicolumn{4}{c}{\textbf{MAE ($\times 10^{-3}$)} ↓} & \multicolumn{3}{c}{\textbf{Forgetting Ratio} ↓} \\
        \midrule
        \multicolumn{1}{c}{\textbf{Experience}} & \textbf{1} & \textbf{2} & \textbf{3} & \textbf{4} & \textbf{1} & \textbf{2} & \textbf{3} \\
        \midrule
        Cumulative & 
        $0.0088 \pm 0.0010$ & $0.0071 \pm 0.0001$ & $0.0078 \pm 0.0004$ & $0.0091 \pm 0.0009$ & 
        $0.19 \pm 0.22$ & $0.11 \pm 0.05$ & $-0.18 \pm 0.06$ \\
        Naive & 
        $0.0523 \pm 0.0002$ & $0.0330 \pm 0.0003$ & $0.0176 \pm 0.0002$ & $0.0070 \pm 0.0002$ & 
        $6.06 \pm 0.88$ & $7.46 \pm 0.08$ & $3.47 \pm 0.10$ \\
        \midrule
        Replay & 
        $\mathbf{0.0107 \pm 0.0010}$ & $0.0092 \pm 0.0005$ & $\mathbf{0.0089 \pm 0.0007}$ & $\mathbf{0.0083 \pm 0.0006}$ & 
        $\mathbf{0.46 \pm 0.22}$ & $0.33 \pm 0.12$ & $0.15 \pm 0.09$ \\
        GEM & 
        $0.0208 \pm 0.0062$ & $\mathbf{0.0058 \pm 0.0023}$ & $0.0129 \pm 0.0051$ & $0.0315 \pm 0.0050$ & 
        $1.82 \pm 1.13$ & $\mathbf{-0.47 \pm 0.24}$ & $\mathbf{-0.34 \pm 0.21}$ \\
        EWC & 
        $0.0470 \pm 0.0009$ & $0.0292 \pm 0.0005$ & $0.0154 \pm 0.0006$ & $0.0073 \pm 0.0001$ & 
        $5.51 \pm 0.79$ & $5.76 \pm 1.11$ & $2.36 \pm 0.13$ \\
        \bottomrule
    \end{tabular}%
    }
    \caption{Comparison of final MAE and Forgetting on each experience across different strategies for the SplitDRIVAERNET++-PC benchmark in the bin incremental scenario}
    \label{tab:comparison_bin_splitdrivaernet++PC}
\end{table*}

Replay achieved the lowest final MPE of 3.64\%, closely matching the Joint strategy’s 3.36\%, and reported the lowest overall forgetting ratio of 0.35. As shown in Table \ref{tab:comparison_splitdrivaernet++PC}, Replay achieved the lowest final MAE on all experiences except experience 2, where GEM slightly outperformed it. Figure \ref{fig:DRIVAERNETplusplus-PC_mae} confirms Replay’s consistent incremental performance, closely tracking the Joint baseline, while Figure \ref{fig:DRIVAERNETplusplus-PC_forgetting} highlights its strong mitigation of forgetting across experiences. Notably, GEM’s forgetting drops below Replay’s at experience 4, indicating an emphasis on predicting well on earlier experiences, but this does not translate into better overall performance in MAE or MPE. GEM showed signs of backward transfer on experiences 2 and 3 but experienced substantial forgetting on experience 1. EWC exhibited the highest forgetting across all experiences, performed similarly to the Naive baseline, and showed steep error increases over time. In terms of efficiency, Replay achieved a 46\% reduction compared to the Joint baseline, while GEM showed minimal improvement at 7\%. EWC reduced runtime by 50\%, although it performed poorly on prediction error. 


\subsection*{Input Incremental Benchmark Outcomes} 
Table \ref{tab:benchmark_comparison_input} compares all strategies based on MPE and average forgetting ratio for the SplitDRIVAERNET++-Par and SplitDRIVAERNET++-PC benchmarks, tested in the input incremental scenario, which are discussed as overall trends. Then, results for both benchmarks are reported, considering both incremental and final performance. We also mention computational efficiency.

\begin{table*}[h]
    \centering
        \small
    \renewcommand{\arraystretch}{0.9}
    \setlength{\aboverulesep}{0.5ex}
    \setlength{\belowrulesep}{0.5ex}
    \begin{tabular}{l *{5}{c} @{\hspace{2em}} *{4}{c}}
        \toprule
        \multicolumn{1}{c}{\textbf{Metrics}} & \multicolumn{5}{c}{\textbf{Mean Percent Error} ↓} & \multicolumn{4}{c}{\textbf{Forgetting Ratio} ↓}\\
        \midrule
        \multicolumn{1}{c}{\textbf{Strategy}} & \textbf{Baseline} & \textbf{Replay} & \textbf{GEM} & \textbf{EWC} & \textbf{Naive} & \textbf{Replay} & \textbf{GEM} & \textbf{EWC} & \textbf{Naive}\\
        \midrule
        SplitDRIVAERNET++-Par & $3.21$ & $\mathbf{3.78}$ & $4.69$ & $4.81$ & $5.00$ & $\mathbf{0.05}$ & $0.56$ & $1.36$ & $1.28$ \\
        SplitDRIVAERNET++-PC & $3.27$ & $\mathbf{3.40}$ & $5.55$ & $5.91$ & $6.04$ & $\mathbf{-0.15}$ & $0.53$ & $ 0.88 $ & $0.88$ \\
        \bottomrule
    \end{tabular}
    \caption{Comparison of Mean Percent Error and Best Forgetting Ratio across different benchmarks and CL methods for the input incremental scenario}
    \label{tab:benchmark_comparison_input}
\end{table*}

\subsubsection*{Overall Trends}

The forgetting experienced in the input incremental scenario was less pronounced than the bin incremental scenario, with the Naive strategies achieving MPEs of 5.00\% and 6.04\% for the SplitDRIVAERNET++-Par and SplitDRIVAERNET++-PC benchmarks, respectively, compared to the MPEs of 10.57\% and 13.22\% in the bin incremental scenario. For most strategies tested in both the parametric and point cloud representations, there was a spike in error at the second experience, followed by a substantial recovery at the third experience. This trend differs from the steady increase typically observed in the bin incremental scenario. The observed behavior likely reflects greater overlap in design features or geometric similarity between the first and third types of cars, suggesting that the second experience introduced a more challenging distribution shift.

Replay consistently showed the greatest resilience, closely approaching cumulative baseline performance in the point cloud representation, and effectively recovering from the initial performance drop in the parametric representation. GEM offered moderate mitigation of forgetting but with significantly less effectiveness compared to Replay, as well as higher computational costs. EWC consistently struggled, closely matching the Naive strategy, highlighting substantial limitations during input distribution shifts.

\paragraph*{SplitDRIVAERNET++-Par}

\begin{table*}[h]
    \centering
    \small
    \renewcommand{\arraystretch}{0.9}
    \setlength{\aboverulesep}{0.5ex}
    \setlength{\belowrulesep}{0.5ex}
    \begin{tabular}{l *{3}{c} @{\hspace{2em}} *{2}{c}}
        \toprule
        \multicolumn{1}{c}{\textbf{Metrics}} & \multicolumn{3}{c}{\textbf{MAE ($\times 10^{-2}$)} ↓} & \multicolumn{2}{c}{\textbf{Forgetting Ratio} ↓} \\
        \midrule
        \multicolumn{1}{c}{\textbf{Experience}} & \textbf{1} & \textbf{2} & \textbf{3} & \textbf{1} & \textbf{2} \\
        \midrule
        Cumulative & 
        $0.966 \pm 0.051$ & $0.736 \pm 0.025$ & $0.716 \pm 0.033$ & 
        $0.26 \pm 0.13$ & $-0.17 \pm 0.06$ \\
        Naive & 
        $2.779 \pm 0.119$ & $0.668 \pm 0.037$ & $0.537 \pm 0.015$ & 
        $2.62 \pm 0.28$ & $0.16 \pm 0.04$ \\
        \midrule
        Replay & 
        $\mathbf{1.336 \pm 0.083}$ & $0.775 \pm 0.026$ & $0.774 \pm 0.028$ & 
        $\mathbf{0.74 \pm 0.21}$ & $\mathbf{-0.40 \pm 0.04}$ \\
        GEM & 
        $1.917 \pm 0.200$ & $0.989 \pm 0.096$ & $0.773 \pm 0.082$ & 
        $1.50 \pm 0.34$ & $-0.08 \pm 0.08$ \\
        EWC & 
        $2.923 \pm 0.138$ & $\mathbf{0.657 \pm 0.065}$ & $\mathbf{0.509 \pm 0.031}$ & 
        $2.81 \pm 0.38$ & $0.13 \pm 0.12$ \\
        \bottomrule
    \end{tabular}%
    \caption{Comparison of final MAE and Forgetting on each experience across different strategies for the SplitDRIVAERNET++-Par benchmark in the input incremental scenario}
    \label{tab:comparison_splitdrivaernet++Par}
\end{table*}

Replay achieved the best MPE of 3.78\%, compared to the Joint MPE of 3.21\%, the lowest overall forgetting across experiences, and the lowest MAE on experience 1, indicating effective resistance to catastrophic forgetting (see Table \ref{tab:comparison_splitdrivaernet++Par}). As shown in Figure \ref{fig:DRIVAERNETplusplus-Par_input_mae}, all strategies experienced error spikes following the introduction of experience 2, with Replay showing notable improvement on experience 2 after training on experience 3. GEM maintained moderate forgetting and MAE performance across experiences, consistently ranking between Replay and EWC. In contrast, EWC achieved the lowest MAEs on the most recent experiences, reflecting high plasticity, but suffered from severe forgetting of earlier tasks—highlighted by a forgetting ratio of 2.81. The overall forgetting trends in Figure \ref{fig:DRIVAERNETplusplus-Par_input_forgetting} further reinforce Replay's advantage in balancing stability and plasticity on this benchmark. Similar to the bin incremental benchmark, Replay demonstrated the largest improvement with a 50\% runtime reduction from the Joint baseline, outperforming both GEM (38\% reduction) and EWC (42\% reduction) while also maintaining better prediction quality and forgetting resistance.

\paragraph*{SplitDRIVAERNET++-PC}
\begin{table*}[h]
    \centering
        \small
    \renewcommand{\arraystretch}{0.9}
    \setlength{\aboverulesep}{0.5ex}
    \setlength{\belowrulesep}{0.5ex}
    \begin{tabular}{l *{3}{c} @{\hspace{2em}} *{2}{c}}
        \toprule
        \multicolumn{1}{c}{\textbf{Metrics}} & \multicolumn{3}{c}{\textbf{MAE ($\times 10^{-2}$)} ↓} & \multicolumn{2}{c}{\textbf{Forgetting Ratio} ↓} \\
        \midrule
        \multicolumn{1}{c}{\textbf{Experience}} & \textbf{1} & \textbf{2} & \textbf{3} & \textbf{1} & \textbf{2} \\
        \midrule
        Cumulative & 
        $0.737 \pm 0.010$ & $0.889 \pm 0.026$ & $0.861 \pm 0.035$ & 
        $-0.21 \pm 0.03$ & $-0.07 \pm 0.06$ \\
        Naive & 
        $2.846 \pm 0.145$ & $0.967 \pm 0.024$ & $0.896 \pm 0.031$ & 
        $2.04 \pm 0.23$ & $-0.03 \pm 0.05$ \\
        \midrule
        Replay & 
        $\mathbf{0.836 \pm 0.054}$ & $\mathbf{0.911 \pm 0.013}$ & $\mathbf{0.893 \pm 0.023}$ & 
        $\mathbf{-0.15 \pm 0.05}$ & $-0.10 \pm 0.02$ \\
        GEM & 
        $2.294 \pm 0.372$ & $1.064 \pm 0.062$ & $0.981 \pm 0.042$ & 
        $1.44 \pm 0.33$ & $\mathbf{-0.13 \pm 0.04}$ \\
        EWC & 
        $2.947 \pm 0.356$ & $0.967 \pm 0.012$ & $0.923 \pm 0.058$ & 
        $2.18 \pm 0.41$ & $-0.02 \pm 0.04$ \\
        \bottomrule
    \end{tabular}%
    \caption{Comparison of final MAE and Forgetting on each experience across different strategies for the SplitDRIVAERNET++-PC benchmark in the input incremental scenario}
    \label{tab:comparison_splitdrivaernet++PC}
\end{table*}

Replay achieved the lowest MPE of 3.40\%, compared to the Joint MPE of 3.27\%, and demonstrated the most consistent performance, achieving the lowest MAE values on all experiences (see Table \ref{tab:comparison_splitdrivaernet++PC}), as well as the lowest overall forgetting (see Table \ref{tab:benchmark_comparison_input}). Figure \ref{fig:DRIVAERNETplusplus-PC_input_mae} confirms that Replay maintained low incremental MAE throughout training, closely tracking the Joint strategy, without spiking in error at experience 2. GEM outperformed EWC, achieving the best forgetting ratio on experience 2, indicating backward transfer and some recovery of prior knowledge. However, GEM still suffered substantial forgetting on experience 1, with a ratio greater than 1. EWC, meanwhile, performed comparably to the Naive baseline, showing severe forgetting across experiences. Figure \ref{fig:DRIVAERNETplusplus-PC_input_forgetting} further illustrates these patterns, with Replay exhibiting minimal forgetting, GEM showing moderate degradation, and EWC experiencing the highest forgetting. The computational efficiency mirrored that of the bin incremental scenario, with Replay reducing runtime by 39\% compared to the Joint baseline, GEM achieving only a 7\% reduction, and EWC showing a 45\% improvement in runtime.


\begin{table*}[!htbp]
\centering
    \small
    \renewcommand{\arraystretch}{0.9}
    \setlength{\aboverulesep}{0.5ex}
    \setlength{\belowrulesep}{0.5ex}
\setlength{\tabcolsep}{6pt}
\renewcommand{\arraystretch}{1}
\begin{tabular}{l l l l l l}
\toprule
    \textbf{Benchmark} & \textbf{Joint} & \textbf{Replay} & \textbf{GEM} & \textbf{EWC} & \textbf{Naive} \\
    \midrule
    SplitSHIPD-Par & 1837.08 & 1408.47 & 1353.75 & \textbf{1169.59} & 990.24 \\
    SplitSHIPD-PC & 20765.72 & 16169.97 & 18487.34 & \textbf{8817.80} & 8342.42 \\

    SplitSHAPENET & 14720.70 & 8154.58 & 13665.93 & \textbf{6852.46} & 5896.23 \\
    SplitRAADL & 1678.76 & 1386.83 & 1245.09 & \textbf{926.54} & 848.12 \\
    SplitDRIVAERNET & 9458.70 & 5038.06 & 8228.63 & \textbf{4593.38} & 3821.70 \\
    SplitDRIVAERNET++-Par (Bin) & 1761.11 & \textbf{718.69} & 1005.57 & 914.91 & 702.50 \\
    SplitDRIVAERNET++-PC (Bin) & 7662.63 & 4150.30 & 8168.27 & \textbf{3796.02} & 3167.49 \\
    SplitDRIVAERNET++-Par (Input) & 1385.63 & \textbf{693.54} & 863.97 & 805.20 & 674.47 \\
    SplitDRIVAERNET++-PC (Input) & 6274.35 & 3801.33 & 5862.19 & \textbf{3428.01} & 3108.58 \\
    \bottomrule
    \end{tabular}
\caption{Total average run time (seconds) across all benchmarks}
\label{tab:timing}
\end{table*}

\begin{figure*}[!tb]
    \includegraphics[width=\textwidth]{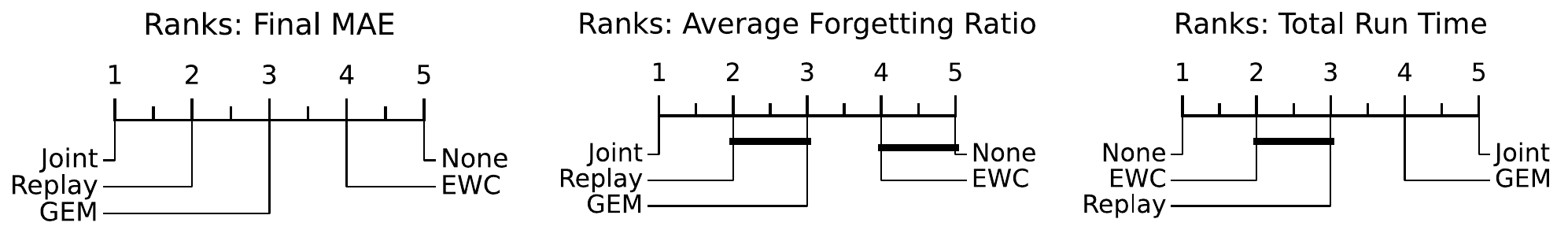}
    \caption{Rank plots of CL strategies for Final MAE, Forgetting Ratio, and  Total Run Time across all nine benchmarks}
    \label{fig:rank_plots}
\end{figure*}

\section{Discussion}

This section discusses the implications of the results obtained by testing Experience Replay (ER), GEM, and EWC in the continual learning setting. We further analyze strategy-specific behaviors observed across the benchmarks, highlighting the conditions under which each strategy showed relative advantages. Then, we emphasize the unique challenges posed by engineering problems in the continual learning context and outline directions for future research.

\subsection*{Statistical Trends Across Benchmarks}
We conducted a statistical analysis of strategy performance across all nine benchmarks using the Mann-Whitney U test on normalized metrics, with results visualized through critical difference (CD) plots. These plots enabled a consistent ranking of strategies across benchmarks with differing scales, covering diverse 3D engineering tasks.

The rank plots for final MAE, forgetting ratio, and total run time can be found in Figure \ref{fig:rank_plots}. Analysis of the normalized forgetting ratios revealed a statistically significant hierarchy ($p < 0.05$). Rehearsal-based strategies, Replay and GEM, ranked just below the ideal Joint baseline and were statistically indistinguishable from one another, as indicated by a shared horizontal bar in the CD plot. However, EWC did not show statistical differences from the Naive baseline in terms of forgetting between benchmarks, with both exhbiting the highest forgetting. For MAE, Joint training achieved the lowest error overall, followed by Replay, GEM, EWC, and finally Naive, with statistically significant differences separating each group. In terms of computational efficiency, EWC and Replay were statistically equivalent in runtime across benchmarks. GEM required considerably more computational resources, while Joint training was the most expensive, as expected.

These conclusions are based on the assumption that our benchmark suite, including variants of SplitSHIPD, SplitSHAPENET, SplitRAADL, and SplitDRIVAERNET, offers representative coverage of 3D engineering problems, and that each dataset contributes equally to the final rankings. Our normalization procedure addresses the wide range of metric scales across benchmarks, making fair statistical comparison possible, though it cannot fully capture the varying importance of individual datasets. Nevertheless, these findings suggest that Replay offers the best trade-off between forgetting resistance, predictive accuracy, and computational cost in the context of continual learning for engineering applications.

\subsection*{Stability vs. Plasticity: Observations on Forgetting}

Analyzing forgetting metrics provided additional insights into the strategies' behaviors. Replay consistently maintained low forgetting across all benchmarks, highlighting its robustness. However, GEM exhibited notably low forgetting on certain benchmarks, sometimes even surpassing the Joint strategy in reducing forgetting. This phenomenon aligns closely with the \textit{stability/plasticity dilemma}, where models that emphasize stability can minimize forgetting at the expense of adaptability. Conversely, overly plastic models may adapt quickly but suffer from higher forgetting. GEM’s strong stability property occasionally resulted in exceptionally low forgetting, especially notable in benchmarks like SHIPD-Parametric and RAADL, however its inability to adapt to new experiences in general reduced its overall performance.

The performance of GEM was also constrained by computational limitations inherent in our experimental setup. In particular, GEM is highly time-intensive when applied to high-dimensional data, such as point clouds, especially with a large number of patterns per experience. To ensure GEM’s total runtime remained shorter than that of the Joint strategy, we had to restrict the number of patterns used to constrain the loss function. This reduction in constraint samples likely contributed to GEM’s diminished overall performance. Such limitations should be taken into account when applying GEM in practice, especially in engineering domains where datasets are often high dimensional.

\subsection*{Unique Aspects of CL for Engineering}
Continual learning in engineering tasks exhibits distinct behaviors compared to traditional classification settings, driven by the prevalence of regression targets and shared structure across experiences. Engineering problems often involve continuous-valued outputs, requiring precise function approximation and fine-grained predictive accuracy. At the same time, many engineering datasets contain underlying similarities across experiences, which can promote positive backward transfer, a phenomenon that is rarely observed in standard CL benchmarks.

We observed this behavior in the SplitRAADL benchmark, where all strategies, even the highly plastic EWC and Naive strategies, improved performance on the first experience after training on later ones. Such negative forgetting highlights the potential for constructive interference between tasks.  We hypothesize this stems from the dataset’s small size and narrow target range, which may enable the model to form generalizable feature representations without significant conflict between experiences.

Furthermore, data representations commonly used in engineering, such as point clouds, meshes, and parametric forms, tended to impact strategy performance. Replay consistently demonstrated strong general performance across both point cloud and parametric data, though its efficacy was slightly diminished in larger datasets and in the parametric representations. GEM’s performance was notably closer to Replay’s when dealing with parametric data, suggesting that representation dimensionality might affect strategy performance, and highlighting the importance of choosing strategies appropriate to the data representation.

\subsection*{Additional Evaluation Metrics}
While our study did not explicitly report the following metrics, we list some of the metrics that could be important in future work.

\paragraph{Memory Efficiency} 
Different CL approaches have varying memory requirements, which can be critical for real-world engineering deployment. Some methods, such as experience replay, require storing past data. Since memory resources are often limited, strategies with lower storage demands are appealing. Tracking the number of parameters or the amount of stored data for each method would help highlight the trade-off between memory efficiency and accuracy.

\paragraph{Forward Transfer and Adaptability}
Beyond mitigating forgetting, an effective CL strategy should also enable positive knowledge transfer to new tasks. This means evaluating whether learning an earlier task improves performance on subsequent ones. For example, in engineering design problems, solving simpler cases first may accelerate learning of more complex ones. Additionally, measuring how quickly a model adapts to each new task, such as tracking the number of iterations needed to reach a certain error threshold, can be crucial. Faster adaptation is especially valuable in iterative design scenarios where efficiency is key.

\paragraph{Real-World Performance Metrics}
Depending on the application, certain real-world performance metrics could be useful to track. One such metric is prediction uncertainty, which would measure whether models maintain their confidence calibration over time. In engineering design optimization, it is also important to ensure that a continually learned model continues to respect physical constraints and design limits established in previous tasks. Assessing these factors using quantifiable metrics would provide a more comprehensive evaluation of CL methods in engineering settings.

\subsection*{Impact and Future Directions}
Overall, the engineering domain presents a compelling and practical application area for continual learning, given its inherently dynamic and data-intensive nature. Retraining from scratch after each data update is often inefficient and increasingly impractical as models and datasets grow in complexity, a trend that is inevitable in engineering workflows. While this work takes an important step toward introducing continual learning into the engineering context, further research is needed to develop scalable, tailored strategies that address the unique aspects of the field.

Our benchmarking results highlight several directions for future research in continual learning for engineering. First, our findings emphasize the potential of data-driven methods like Experience Replay, which consistently achieved strong performance by directly leveraging the training data pipeline. However, improving replay efficiency, particularly for larger and more diverse datasets, remains an important challenge for scaling Replay effectively. In contrast, theoretically motivated approaches such as GEM and EWC, which aim to mitigate forgetting through loss function regularization, showed notable limitations. GEM often incurred high computational cost, approaching that of full retraining, while still suffering from performance degradation. EWC was computationally efficient but yielded only marginal improvements over the Naive baseline. These observations suggest that future work should prioritize scalable, data-driven strategies that better balance predictive performance and computational efficiency.

Finally, our study did not evaluate multi-target incremental scenarios involving architectural strategies. Investigating whether dynamic architectures can handle incremental learning in engineering problems would provide valuable insights for optimizing continual learning performance. Understanding these trade-offs could significantly advance the field and enhance practical applicability.

\section{Conclusions}
This paper introduced continual learning strategies to engineering regression tasks, particularly surrogate modeling for computationally intensive simulations. It demonstrated the effectiveness of adapted CL methods, notably Experience Replay, in mitigating catastrophic forgetting for complex 3D engineering datasets. By proposing novel regression-specific CL scenarios and benchmarks, the study provided practical resources tailored to engineering needs. Results highlighted key factors influencing CL performance, such as dataset size, dimensionality, and data representation. Future research should further develop specialized CL strategies addressing the unique challenges of engineering design problems.

\begin{acknowledgment}
We thank Matthew Jones at the MIT Lincoln Laboratory for providing access to the RAADL glider dataset used in this study and for helpful discussions regarding its use. This work would not have been possible without their contribution. 
This research was conducted as part of the DeCode Lab at MIT. The work was supported in part by the U.S. Air Force through the DAWN-ED scholarship program. The views expressed are those of the authors and do not necessarily reflect the official policy or position of the Department of the Air Force.
\end{acknowledgment}

\bibliographystyle{asmems4}

\bibliography{asme2e}

\appendix       
\section*{Appendix A: Detailed Dataset Descriptions}

\paragraph{SHIPD} The SHIPD dataset \cite{bagazinski2023ship, DVN/MMGAUS_2024}  was originally introduced in 2023 to support the application of ML methods to naval architecture design. The dataset contains 30,000 sample ship hull designs and their associated wave drag coefficients, consisting of three constrained sets of 10,000 samples. The designs are represented as point clouds of the ship hulls, capturing the geometry with 20,000 points along the surface. We use the first constrained set of 10,000 samples in our experiments, as the computational cost of training point cloud-based models increases significantly with dataset size. Due to the high resolution of the geometry (20,000 points per hull) and the complexity of the models, training times scale considerably, making the full dataset impractical for thorough experimentation within our resource and time constraints. The same 10,000 designs are also offered as sets of 44 parameters that define the shape of each hull, which we also include in our benchmarking study. This dataset was chosen for its connection to surrogate modeling applications in engineering for naval applications. Furthermore, as the largest dataset out of the five datasets tested and one of two datasets that offer both parametric and point cloud data, it provides a path for understanding how certain CL strategies may respond to larger datasets and different data representations.

 \paragraph{DrivAerNet and DrivAerNet++} The DrivAerNet \cite{elrefaie2025drivaernet} and DrivAerNet++ \cite{elrefaie2025drivaernet++} datasets, released in 2024 and 2025, respectively, provide car geometries and their associated drag values, obtained through high-fidelity CFD simulations. DrivAerNet includes 4,000 car designs in the Fastback category offered as meshes and point clouds, and their associated drag coefficients. DrivAerNet++ added an additional 4,000 designs, including designs from the Fastback, Notchback, and Estateback categories. The additional 4,000 car designs are offered as both parametric data and point clouds, so we use this selection of the data to run experiments for both representations. The DrivAerNet and DrivAerNet++ datasets are among the only aerodynamic datasets that model the wheels and underbody of cars, and they provide several design parameters for their geometric representation \cite{elrefaie2025drivaernet}. We include this dataset as an aerodynamics example of the surrogate modeling problem with detailed car designs.

 \paragraph{ShapeNet Car} The ShapeNet Car dataset \cite{song2023surrogate} was released in 2023, accompanied by a surrogate model for predicting drag using 2D representations of the data. The dataset contains 9,070 car designs as meshes, along with their associated drag coefficients. In this study, the meshes are converted to 3D point clouds to be used for benchmarking. We include this dataset as another application in aerodynamics for cars, providing a different modeling and parameterization than the DrivAerNet dataset.
 
\paragraph{RAADL} The Rapid Aerodynamic and Deep Learning (RAADL) dataset was provided by MIT Lincoln Laboratory \cite{jones2024rapid}. It consists of 800 glider geometries and their associated drag coefficients, lift coefficients, and moment coefficients, obtained through CFD simulations. The gliders are offered as 3D voxelized geometries as well as 3D meshes. We convert the meshes to 3D point clouds and use the drag coefficients as targets. We include this dataset to provide an aerospace application in aircraft design. It also serves as an example of a small engineering dataset, providing insight into how size affects the continual learning problem.

\end{document}